\pdfoutput=1
\documentclass[11pt]{article}

\usepackage[preprint]{acl}
\usepackage{times}
\usepackage{latexsym}
\usepackage[T1]{fontenc}
\usepackage[utf8]{inputenc}
\usepackage{microtype}
\usepackage{inconsolata}
\usepackage{booktabs}
\usepackage{graphicx}
\usepackage{svg}
\graphicspath{{figures/}}
\usepackage{xcolor}
\usepackage{amsmath}
\usepackage{amsfonts}
\usepackage{url}
\usepackage{hyperref}
\usepackage[breakable]{tcolorbox}
\usepackage{listings}
\usepackage{enumitem}
\lstdefinestyle{prompt}{
  basicstyle=\scriptsize,
  breaklines=true,
  breakatwhitespace=true,
  columns=fullflexible,
  keepspaces=true,
  showstringspaces=false,
  upquote=true,
  extendedchars=true,
  inputencoding=utf8,
  literate=%
    {`}{{\textasciigrave}}1
    {→}{$\rightarrow$}1
    {←}{$\leftarrow$}1
    {↔}{$\leftrightarrow$}1
    {⇒}{$\Rightarrow$}1
    {—}{---}1
    {–}{--}1
    {…}{\ldots}1
    {≥}{$\geq$}1
    {≤}{$\leq$}1
    {≠}{$\neq$}1
    {×}{$\times$}1
    {±}{$\pm$}1
    {•}{\textbullet}1
    {α}{$\alpha$}1
    {β}{$\beta$}1
    {μ}{$\mu$}1
    {’}{'}1
    {‘}{`}1
    {“}{``}1
    {”}{''}1
    {é}{\'{e}}1
    {è}{\`{e}}1
    {ê}{\^{e}}1
    {à}{\`{a}}1
    {ç}{\c{c}}1
    {ô}{\^{o}}1
    {î}{\^{i}}1
    {ï}{\"{\i}}1
    {ù}{\`{u}}1
    {û}{\^{u}}1
    {œ}{\oe{}}1
    {Œ}{\OE{}}1
    {É}{\'{E}}1
    {À}{\`{A}}1
}
\newcounter{prompt}
\renewcommand{\theprompt}{\arabic{prompt}}
\newcounter{rephraseexample}
\renewcommand{\therephraseexample}{\arabic{rephraseexample}}

\title{Where Does the Signal Live? \\ A Web Data Recipe for Medical Encoder Pretraining}

\author{
  \textbf{Bofeng Huang} \quad \textbf{Jacques Sun}\thanks{\,Work done while at Doctolib.} \\
  \textbf{Diane Bouchacourt} \quad \textbf{Nicolas Barascud}\thanks{\,Equal contribution.} \quad \textbf{Fajwel Fogel}\footnotemark[\value{footnote}] \\[4pt]
  {\normalfont Doctolib} \\
  {\normalfont\texttt{firstname.lastname@doctolib.com}}
}

\begin{document}
\maketitle

\begin{abstract}
Web data curation has been widely studied for decoder Large Language Model (LLM) pretraining.
Encoders for dense-terminology domains such as medicine, by contrast, are pretrained on small, manually-curated corpora that limit scalability and writing style diversity, a bottleneck even more severe in non-English clinical settings. Whether web-scale data curation also benefits encoder Masked Language Modeling (MLM) in a dense-terminology domain remains an open question. To address this, we introduce two complementary levers.
\emph{Medical-term density filtering} selects documents rich in medical terms.
\emph{Signal-amplifying rephrasing} uses an LLM to rewrite documents into denser variants with broader entity contexts.
We instantiate the recipe on French medical NLP. The medical-term density filter outperforms the widely-used educational quality filter on downstream medical tasks, and the two complement each other. Signal-amplifying rephrasing alone improves on raw web data, and mixing it with filtered web data produces the largest gain.
The recipe yields \emph{FineMed}, a French medical pretraining corpus, and \emph{DoctoBERT}, a state-of-the-art French medical encoder family evaluated on both the public benchmark DrBenchmark and a proprietary clinical Named Entity Recognition (NER) task.
\end{abstract}

\section{Introduction}
\label{sec:intro}

Web data curation is widely adopted for decoder Large Language Model (LLM) pretraining: model-based filtering selects documents with signals like educational quality, a scorer of documents' value for student learning~\citep{li_datacomplm_2025, lozhkov2024fineweb-edu, su_nemotroncc_2025}.
Large-scale rephrasing further raises token utility~\citep{hao_reformulation_2025, maini_rephrasing_2024, team_kimi_2026, yu_repro_2025}.
Medical encoders, however, draw from a narrow set of medical sources (e.g., biomedical literature and clinical narratives) assembled manually from a small number of canonical repositories at substantial human cost~\citep{gu_domainspecific_2022, labrak_drbert_2023, lee_biobert_2020, sounack_bioclinical_2025, touchent_camembertbio_2024}.
This sourcing pattern restricts corpus scalability, source heterogeneity, and register diversity. Whether web-scale data curation extends to encoder Masked Language Modeling (MLM) in a dense-terminology domain is largely unstudied.

We focus on two properties of encoder MLM that standard curation often overlooks: per-token entity density~\citep{levine_pmimasking_2020}, where rare or domain-specific tokens yield larger gradients per masked position, and per-entity context diversity, where varied co-occurrence contexts around each entity strengthen its learned representation.
Common LLM data curation, such as educational-quality filtering, rewards documents valuable for student learning, since such documents tend to be coherent prose with lay explanations that dilute specialized vocabulary.
Massive Genre--Audience (MGA)~\citep{hao_reformulation_2025} rephrasing diversifies (genre, audience) framing across documents but does not target domain-specific terminology.
We address both gaps with two complementary levers: \emph{medical-term density filtering}, a per-document filter on medical-term richness, and \emph{signal-amplifying rephrasing}, an LLM rewriter that produces denser variants with varied entity contexts.

We apply this recipe to French medical NLP, where data scarcity is a bottleneck and evaluation benchmarks are well-established. Existing French medical encoders often rely on narrow, manually-curated corpora~\citep{knafou_transbert_2025, labrak_drbert_2023}. We filter French medical content from three general-purpose, heterogeneous web corpora: FineWeb-2~\citep{penedo_fineweb2_2025}, FinePDFs~\citep{kydlicek2025finepdfs}, and FineWiki~\citep{penedo2025finewiki}. We annotate each retained document along three axes (subdomain, educational quality, medical-term density) and ablate filtering and rephrasing recipes against these baseline corpora.
Our ablations show that medical-term density beats educational quality as a single-axis filter. The two combined improve further, and adding signal-amplifying rephrasing on top of filtered raw data extends the gain.
The combined recipe produces \emph{DoctoBERT}, French medical encoders that achieve state-of-the-art performance on DrBenchmark and a proprietary clinical Named Entity Recognition (NER) task from a real-world production setting. Our contributions are four-fold:
\begin{itemize}[noitemsep,topsep=0pt,partopsep=0pt,parsep=0pt]
  \item We propose \emph{medical-term density filtering}, a per-document filter on medical-term richness based on extracted medical entities (\S\ref{sec:medterm}, \S\ref{sec:exp-filtering}).
  \item We introduce \emph{signal-amplifying rephrasing}, a medical adaptation of MGA that raises entity density and broadens entity contexts (\S\ref{sec:rephrasing}, \S\ref{sec:exp-rephrasing}).
  \item We release \emph{FineMed}\footnote{\url{https://huggingface.co/collections/doctolib-lab/finemed-fr}}, a large-scale French medical pretraining corpus annotated along three axes, together with \emph{FineMed-rephrased}, and a reproducible curation pipeline\footnote{\url{https://github.com/doctolib-lab/doctobert}} with multi-axis annotators.
  \item We release \emph{DoctoBERT}\footnote{\url{https://huggingface.co/collections/doctolib-lab/doctobert-fr}}, a family of French medical encoders that achieve state-of-the-art performance on DrBenchmark and a proprietary clinical NER task.
\end{itemize}

\section{Related Work}
\label{sec:related}

\paragraph{Web data curation.}
Modern web pretraining corpora apply heuristic filters and deduplication as standard preprocessing~\citep{gao_pile_2020, penedo_refinedweb_2023, penedo_fineweb_2024, weber_redpajama_2024}.
Beyond this baseline, model-based filtering uses a learned classifier to score documents on quality or domain relevance~\citep{li_datacomplm_2025, lozhkov2024fineweb-edu, wettig_organize_2025}, with educational quality as a particularly important signal.
Most of these studies target decoder LLM next-token loss.
Our results show that for encoder MLM in a dense-terminology domain, per-token entity density outperforms educational quality as a filtering signal.

\paragraph{LLM rephrasing.}
LLM rephrasing of source documents covers creative reformulation across styles and audiences (e.g., MGA~\citep{hao_reformulation_2025})~\citep{maini_rephrasing_2024, niklaus_how_2026} and faithful, constrained edits to prevent hallucinated content from corrupting pretraining~\citep{bi_refinex_2025, yu_repro_2025, zhou_programming_2025}.
Recent work scales the joint filter-plus-rephrase recipe to trillions of tokens~\citep{datologyai_beyondweb_2025, su_nemotroncc_2025}. Mixing rephrased and natural web text outperforms either alone~\citep{kang_demystifying_2025}.
LLM rephrasing largely targets decoder LLM pretraining, which is typically single-epoch at scale~\citep{brown_language_2020, hernandez_scaling_2022, muennighoff_scaling_2025}.
For encoder MLM in a dense-terminology domain, multiple epochs over the same corpus are standard~\citep{devlin_bert_2019}. Hallucinated content corrupts the training distribution across epochs, so our recipe applies a stricter entity-faithfulness constraint.

\paragraph{Medical encoders.}
Domain encoders are typically obtained via continual pretraining of a general encoder~\citep{alsentzer_publicly_2019, lee_biobert_2020, lee_clinical_2025, peng_transfer_2019, sounack_bioclinical_2025} or from-scratch in-domain pretraining with an adapted vocabulary~\citep{beltagy_scibert_2019, fang_bioformer_2023, gu_domainspecific_2022}.
Both depend on a narrow set of curated in-domain corpora, labor-intensive to assemble.
French medical encoders follow the same pattern. Their corpora are scraped from a few canonical websites~\citep{berhe_alibert_2023, labrak_drbert_2023, touchent_camembertbio_2024, touchent_causal_2026}, translated from English medical sources such as PubMed~\citep{knafou_transbert_2025}, or synthesized~\citep{tannier_parhaf_2026}.
Our approach filters and rephrases heterogeneous web data to target encoder MLM.
Recent multilingual web corpora~\citep{kydlicek2025finepdfs, penedo_fineweb2_2025, penedo2025finewiki} have made this practical, but to our knowledge, this methodology has not been applied to French medical NLP.

\section{Methodology}
\label{sec:method}
\begin{figure*}[t]
\centering
\includegraphics[width=\linewidth]{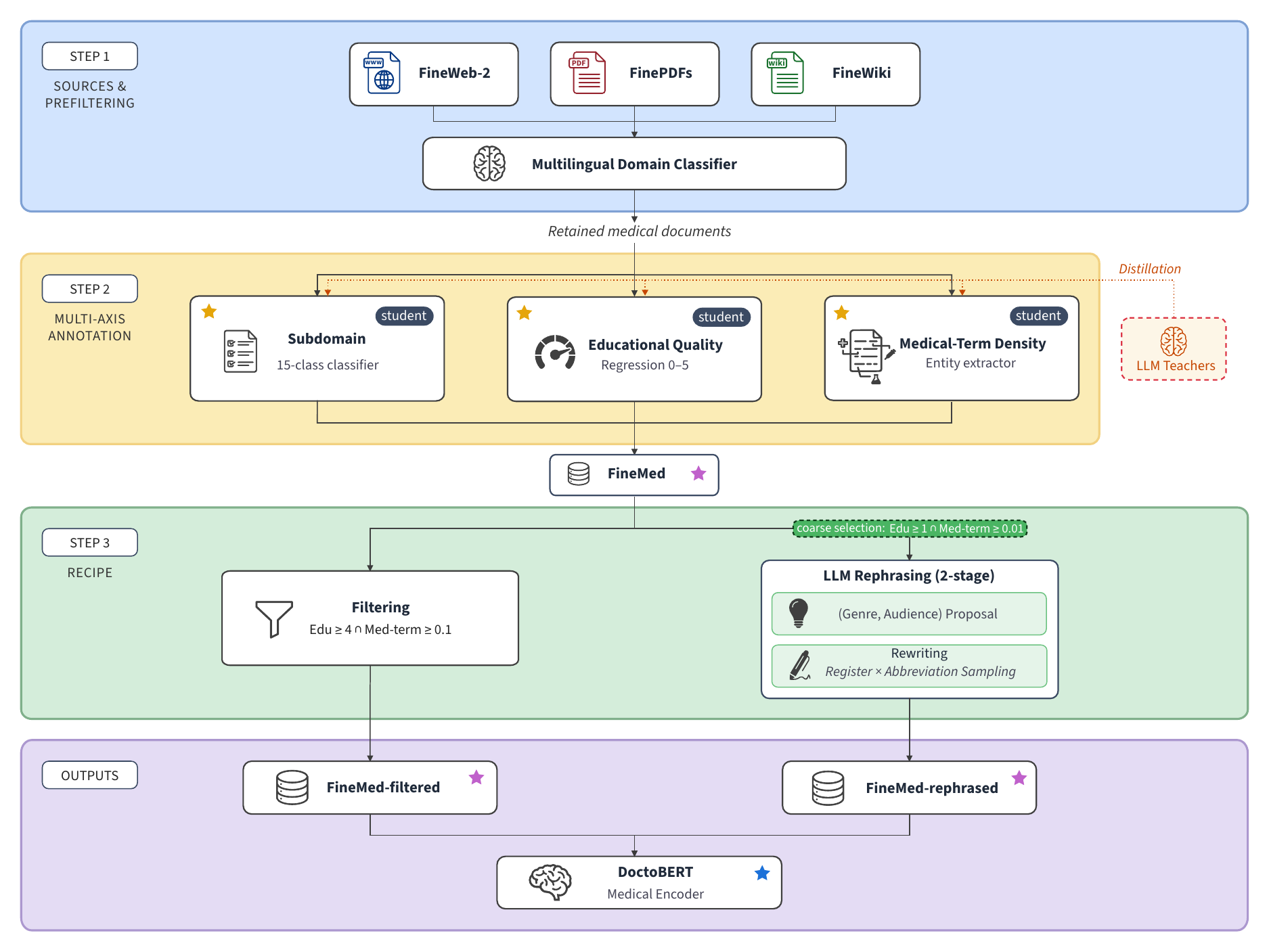}
\caption{Pipeline overview. \textbf{Step 1.} Medical-content prefiltering retains medical
documents from FineWeb-2, FinePDFs, and FineWiki via a multilingual domain
classifier. \textbf{Step 2.} Three small annotators, distilled from LLM teachers,
score each retained document along a different axis: subdomain (15-class
classifier), educational quality (0--5 regression scorer), and medical-term
density (entity extractor). The annotated retained-medical corpus is released
as \emph{FineMed}, unfiltered for task-specific selection downstream. \textbf{Step 3.}
From \emph{FineMed}, the recipe derives \emph{FineMed-filtered} via thresholding on the
multi-axis annotations, and
\emph{FineMed-rephrased} via 2-stage LLM rephrasing process, gated by a coarse-selection
filter. \emph{DoctoBERT}, our medical
encoder family, is pretrained on the mixture of \emph{FineMed-filtered} and \emph{FineMed-rephrased}.
Star colors group released artifacts by type:
\textcolor{yellow}{$\star$} distilled annotators,
\textcolor{violet}{$\star$} datasets (\emph{FineMed}, \emph{FineMed-filtered}, \emph{FineMed-rephrased}),
\textcolor{blue}{$\star$} medical encoders (\emph{DoctoBERT}).}
\label{fig:pipeline}
\end{figure*}

We employ a three-stage pipeline to curate medical pretraining data from the web (Figure~\ref{fig:pipeline}).
A prefiltering step (\S\ref{sec:prefilter}) first isolates documents with actual medical content from the surrounding noise.
The retained documents are then scored along three axes by a multi-axis annotator (\S\ref{sec:annotation}): subdomain, educational quality, and \emph{medical-term density}, the filter signal we introduce.
\emph{Signal-amplifying rephrasing} (\S\ref{sec:rephrasing}), our medical adaptation of MGA, densifies the learning signal at the token level.
In \S\ref{sec:exp} we ablate filter axes, rephrasing recipes, and their combinations.
\S\ref{sec:final} then applies the validated recipe at full scale, training \emph{DoctoBERT} on a corpus that combines filtered and rephrased data.

\subsection{Sources and Prefiltering}
\label{sec:prefilter}

We draw from three large-scale heterogeneous web corpora: FineWeb-2, FinePDFs, and FineWiki.
These provide the scale, source heterogeneity, and stylistic range that curated medical corpora often lack.
The three corpora have been processed through standard LLM pretraining curation (e.g., language identification, heuristic quality filtering, and deduplication), which we inherit from them as a quality baseline.

\textit{Medical prefiltering.}
Medical content only represents a small fraction of each source, which is further diluted by commercial pages. To retain only the actual medical content (Step 1 of Figure~\ref{fig:pipeline}), we apply a pretrained multilingual domain classifier~\citep{nvidia_multilingual_domain_classifier}, reducing each source to under 10\% of its raw size. Per-source retention figures and domain distributions are in Appendix~\ref{app:source-data}.

\subsection{Multi-Axis Annotation}
\label{sec:annotation}

Prefiltering retains every document classified as medical, though their relevance to medical-encoder pretraining differs along multiple dimensions that a single binary label cannot capture. We therefore annotate each retained document (Step 2 of Figure~\ref{fig:pipeline}) along three complementary axes (subdomain, educational quality, and medical-term density), designed to be composable: \S\ref{sec:exp-filtering} reports how thresholds and combinations are selected.

\textit{Annotator distillation.}
Annotating the full corpus with an LLM is prohibitively expensive.
Following Organize-the-Web~\citep{wettig_organize_2025}, we distill teacher LLMs into a lightweight annotator per axis, which drops corpus-level inference cost by an order of magnitude (Appendix~\ref{app:axis-compute}).

\subsubsection{Medical Subdomain}
\label{sec:subdomain}

Medical web content mixes biomedical and clinical writing (e.g., scientific papers, clinical guidelines) with consumer-facing material (e.g., wellness blogs, commercial health pages). To target topics relevant to medical-encoder pretraining, we annotate each retained document with a granular medical-subdomain label.

\textit{Taxonomy.}
Through iterative LLM annotation and human review, we converge on a 15-class medical-subdomain taxonomy that balances coverage and per-class separability (taxonomy in Appendix~\ref{app:subdomain-classifier}).

\textit{Subdomain classifier.}
We fine-tune a ModernCamemBERT~\citep{antoun_modernbert_2025} classifier under a two-stage schedule: a smaller LLM teacher provides high-volume supervision, and a larger LLM teacher provides high-quality supervision.
We apply the classifier across the full corpus, with content and URL as input.

\subsubsection{Educational Quality}
\label{sec:edu}

Subdomain captures what a given document is about, but not how \textit{instructive} it is. A scientific review and a promotional blog post about the same supplement can share the same subdomain while differing strongly in their value for medical training. We therefore score each document on educational quality, a 0--5 score adapted from FineWeb-Edu's general-education rubric~\citep{lozhkov2024fineweb-edu}.

\textit{Scoring rules.}
We adapt FineWeb-Edu's additive 0--5 scoring from general school education to medical education through iterative LLM annotation and human review (scoring rules in Appendix~\ref{app:edu-classifier}).

\textit{Educational-quality scorer.}
We fine-tune a ModernCamemBERT regression scorer under the same two-stage schedule as in \S\ref{sec:subdomain}.
We apply the scorer across the full corpus, with document content as input.

\subsubsection{Medical-Term Density}
\label{sec:medterm}

A document with relevant subdomain and high educational quality can still be sparse in medical terminology. For encoder MLM, most masked tokens in sparse documents fall on non-medical text, so the encoder learns less medical content per pass. We propose \emph{medical-term density} to measure the per-document concentration of medical terms.

\textit{Definition.}
Density is the ratio of characters in extracted medical-term spans to total characters in a document:
\begin{equation}
\text{density} = \frac{\text{\# medical-term characters}}{\text{\# total characters}}.
\end{equation}
We use characters rather than words because medical terms can span multiple subword tokens, and character counts approximate MLM's token-level masking more closely.
Unlike a regression model that may not generalize across document lengths and formats, this character-ratio definition is more robust and exposes the spans for inspection.

\textit{Medical entity extractor.}
We fine-tune GLiNER2~\citep{zaratiana2025gliner2efficientmultitaskinformation} on LLM annotations to identify medical-entity spans.
These spans follow an 8-class taxonomy that we adapt from UMLS (Unified Medical Language System) entity groups to focus on medical-term-rich classes (full taxonomy in Appendix~\ref{app:medterm-extractor}).
The three axes are correlated but capture distinct signals (Appendix~\ref{app:joint-dist}), motivating the joint-filter ablation in \S\ref{sec:exp-filtering}.

\subsection{Signal-Amplifying Rephrasing}
\label{sec:rephrasing}

Filtering with these multi-axis annotations selects high-scoring documents, but cannot enrich the medical content within them. It also discards borderline documents that, despite falling short on one axis, still contain non-negligible medical content worth recovering.
To amplify signal within retained documents and recover it from discarded ones, we use an LLM to rephrase each document into a faithful variant (Step 3 of Figure~\ref{fig:pipeline}) that raises medical-term density and broadens the co-occurrence context around each medical concept.
This matters especially for encoder MLM pretraining, where multiple epochs over the same corpus compound the per-document gain, unlike decoder LLM's typical single-epoch regime.
Rephrasing also cleans up source-level artifacts that filtering leaves untouched (e.g., FineWeb-2 boilerplate, FinePDFs OCR errors).

We adapt Massive Genre--Audience reformulation (MGA)~\citep{hao_reformulation_2025}, a two-stage LLM-rephrasing recipe which provides control over corpus-level style diversity.
Stage~1 plans diverse rephrasings by proposing (genre, audience) pairs for a source document, and stage~2 executes each rephrasing for its assigned pair (e.g., a Wikipedia drug entry rephrased as a pharmacist's reference).
The following paragraphs describe our medical adaptations to each stage (full prompts in Appendix~\ref{app:rephrasing-prompts}).

\textit{Medical-content gating.}
Even after the \S\ref{sec:prefilter} prefilter, some retained documents contain insufficient medical content for effective rephrasing (for example, commercial product pages or general wellness articles that mention medical terms only incidentally).
We therefore add a gating step at the start of stage~1: the LLM assesses whether a document's medical content is sufficient, and failed documents are discarded before stage~2.
This reduces computational overhead and prevents the LLM from fabricating medical content.

\textit{Diverse pair proposals.}
Standard MGA stage~1 generates (genre, audience) pairs jointly, which can default to a small set of repeated pairs across documents, especially in narrow domains like medical.
To broaden coverage, we generate multiple candidate pairs per document and sample one for rephrasing (one variant per source).
To build this pool, the LLM proposes candidate genres and audiences independently, then couples them to ensure real-world plausibility and broad diversity.

\textit{Faithful densification.}
Stage~2 preserves medical content while stripping non-medical filler. This raises per-token medical density and keeps the surrounding context needed for MLM.
Importantly, we instruct the LLM to be strictly meaning-preserving: no medical facts, values, or entities should be invented (a failure mode of unconstrained LLM rephrasing we guard against).
Non-medical Personally Identifiable Information (PII; e.g., names, addresses) is the one exception: the LLM replaces it with varied fictional values, supporting downstream de-identification robustness~\citep{sounack_bioclinical_2025}.

\textit{Medical surface variation.}
Beyond the assigned (genre, audience) pair, stage~2 also varies along two dimensions: register (formal or telegraphic, i.e., clinical-notes style) and abbreviation density (expanded, moderate, or heavy). These dimensions target medical-text style variations (clinical notes vs.\ patient education; abbreviation-heavy vs.\ spelled-out terminology) and broaden each entity's contextual coverage beyond the (genre, audience) pair alone.

\section{Experiments}
\label{sec:exp}

To identify the best data curation recipe, we run filtering and rephrasing ablations at matched training-token compute (Appendix~\ref{app:pretraining}).
For each ablation, we pretrain a ModernBERT~\citep{warner_smarter_2024} encoder from scratch on a candidate corpus, then evaluate on DrBenchmark~\citep{labrak_drbenchmark_2024}.
Candidate corpora are filtering or rephrasing variants of FW2-Med (a medical-prefiltered subset of FineWeb-2).
The filtering ablation (\S\ref{sec:exp-filtering}) tests the three annotated axes (\S\ref{sec:annotation}) at multiple thresholds where applicable, alone and combined.
The rephrasing ablation (\S\ref{sec:exp-rephrasing}) compares rephrasing recipes (\S\ref{sec:rephrasing}), alone and mixed with the best filtered raw data.

\textit{Evaluation.}
We adapt DrBenchmark\footnote{\url{https://github.com/doctolib-lab/DrBenchmark}}, a French medical NLP benchmark covering biomedical and clinical tasks, by replacing fixed hyperparameters with Hyperparameter Optimization (HPO) on the validation split and dropping noisy tasks, resulting in a 7-task subset (Appendix~\ref{app:adapted-benchmark}).
The retained tasks span biomedical NER (QUAERO EMEA/MEDLINE), clinical NER (E3C Clinical/Temporal, DEFT2021), biomedical specialty classification (MORFITT), and clinical diagnostic-category classification (DIAMED).
Cross-task aggregation reports \emph{Min-Max normalized scores} (magnitude-sensitive) and pairwise \emph{Win Probability} (rank-based) to produce a per-model score, avoiding the bias of plain averaging across tasks of different scales.
\subsection{Filtering: Single Axes and Combinations}
\label{sec:exp-filtering}

\begin{table}[t]
\centering
\footnotesize
\setlength{\tabcolsep}{3pt}
\begin{tabular}{l r@{}l r@{}l}
\toprule
Configuration & \multicolumn{2}{c}{Min-Max} & \multicolumn{2}{c}{WP} \\
\midrule
\multicolumn{5}{l}{\emph{External baselines}} \\
NACHOS                          & 45.57 & ${}_{\pm 13.52}$ & 40.26 & ${}_{\pm 6.04}$ \\
TransCorpus-bio-fr              & 38.41 & ${}_{\pm 14.71}$ & 32.47 & ${}_{\pm 4.75}$ \\
\midrule
\multicolumn{5}{l}{\emph{FW2-Med}} \\
Unfiltered                       & 45.02 & ${}_{\pm 15.47}$ & 45.45 & ${}_{\pm 3.76}$ \\
\midrule
\multicolumn{5}{l}{\emph{FW2-Med: single-axis filters}} \\
Bio\&Cli                                & 63.59 & ${}_{\pm 13.27}$ & 50.65 & ${}_{\pm 2.96}$ \\
Edu $\geq 2$                            & 47.50 & ${}_{\pm 10.40}$ & 31.17 & ${}_{\pm 5.39}$ \\
Edu $\geq 4$                            & 59.16 & ${}_{\pm 10.10}$ & 40.26 & ${}_{\pm 6.04}$ \\
Med-term $\geq 0.1$                     & \underline{78.27} & ${}_{\pm 6.21}$ & \underline{66.23} & ${}_{\pm 4.82}$ \\
Med-term $\geq 0.2$                     & 71.28 & ${}_{\pm 11.95}$ & 58.44 & ${}_{\pm 4.89}$ \\
\midrule
\multicolumn{5}{l}{\emph{FW2-Med: combinations}} \\
Bio\&Cli $\cap$ Edu $\geq 4$            & 62.64 & ${}_{\pm 8.66}$ & 46.75 & ${}_{\pm 4.75}$ \\
Bio\&Cli $\cap$ Med-term $\geq 0.1$     & 77.40 & ${}_{\pm 5.83}$ & \underline{66.23} & ${}_{\pm 6.47}$ \\
Bio\&Cli $\cap$ Med-term $\geq 0.2$     & 61.16 & ${}_{\pm 15.75}$ & 53.25 & ${}_{\pm 4.35}$ \\
Edu $\geq 4$ $\cap$ Med-term $\geq 0.1$ & \textbf{81.25} & ${}_{\pm 4.02}$ & \textbf{68.83} & ${}_{\pm 5.39}$ \\
\bottomrule
\end{tabular}
\caption{Single- and multi-axis filtering ablation. The intersection of educational quality (Edu) and medical-term density (Med-term) outperforms both single-axis filters and curated medical corpora. Min-Max (normalized per task) and Win Probability (WP) aggregate scores across tasks, capturing relative magnitude and consistency respectively, both on a 0--100 scale. Values are mean$\pm$SE; best per metric in bold, second underlined. Per-task scores in Appendix~\ref{app:per-task}.}
\label{tab:filter-ablation}
\end{table}

At matched compute, we apply each filter to FW2-Med and pretrain for 20B tokens per configuration.
The subdomain filter restricts to the biomedical and clinical classes (Bio\&Cli, Appendix~\ref{app:bio-cli}), our target subdomains for medical-encoder pretraining.
For Edu and Med-term, we test several thresholds per axis.
As external baselines, we use the pretraining corpora of two existing French medical encoders: NACHOS (used to pretrain DrBERT~\citep{labrak_drbert_2023}) and TransCorpus-bio-fr (translated PubMed, used to pretrain TransBERT-bio-fr~\citep{knafou_transbert_2025}).
Table~\ref{tab:filter-ablation} reports cross-task scores for each configuration.

\paragraph{Web data is competitive, and medical-term density is the strongest single-axis filter.}
Unfiltered FW2-Med ranks alongside NACHOS and above TransCorpus-bio-fr: web data is competitive with curated medical corpora.
Web sources span wider registers, formats, and topics than narrowly sourced curated corpora, and FineWeb-2 inherits modern text-quality curation.
Among single-axis filters, medical-term density is strongest, subdomain second, educational quality weakest.
For decoder LLMs, educational quality is a dominant filter signal; for encoder MLM in a dense-terminology domain, signals targeting domain-token concentration carry more weight.

\paragraph{Combining educational quality and medical-term density beats either alone.}
The intersection of educational quality and medical-term density is the strongest filter: it outperforms medical-term density alone and beats external baselines by +29 WP points.
Educational quality rewards coherent, well-structured prose; medical-term density rewards terminology richness; the two compose.
\subsection{Rephrasing: Recipes and Mixes}
\label{sec:exp-rephrasing}

\begin{table}[t]
\centering
\footnotesize
\setlength{\tabcolsep}{3pt}
\begin{tabular}{l r@{}l r@{}l}
\toprule
Configuration & \multicolumn{2}{c}{Min-Max} & \multicolumn{2}{c}{WP} \\
\midrule
\multicolumn{5}{l}{\emph{FW2-Med}} \\
No rephrasing                           & 50.36 & ${}_{\pm 7.29}$ & 54.76 & ${}_{\pm 20.34}$ \\
\midrule
\multicolumn{5}{l}{\emph{Standard MGA}} \\
Qwen3.5-35B-A3B                         & 27.50 & ${}_{\pm 11.08}$ & 26.19 & ${}_{\pm 13.51}$ \\
\midrule
\multicolumn{5}{l}{\emph{Our recipe, varying the rephraser}} \\
Qwen3.5-35B-A3B                         & \textbf{97.56} & ${}_{\pm 2.44}$ & \textbf{95.24} & ${}_{\pm 3.01}$ \\
Qwen3.5-122B-A10B                       & \underline{89.30} & ${}_{\pm 2.95}$ & \underline{73.81} & ${}_{\pm 13.51}$ \\
Gemma-4-26B-A4B                         & 81.67 & ${}_{\pm 5.32}$ & 71.43 & ${}_{\pm 14.29}$ \\
MedGemma-27B                            & 22.22 & ${}_{\pm 6.33}$ & 23.81 & ${}_{\pm 15.50}$ \\
GPT-OSS-120B                            & 2.35 & ${}_{\pm 2.35}$ & 4.76 & ${}_{\pm 3.01}$ \\
\bottomrule
\end{tabular}
\caption{Rephrasing ablation at the 100k-source-document scale. Our recipe outperforms unrephrased FW2-Med, while standard MGA falls below it. Metrics and formatting as in Table~\ref{tab:filter-ablation}. Per-task scores in Table~\ref{tab:rephrase-100k-per-task}.}
\label{tab:rephrase-100k}
\end{table}

\begin{table}[t]
\centering
\footnotesize
\setlength{\tabcolsep}{3pt}
\begin{tabular}{l r@{}l r@{}l}
\toprule
Configuration & \multicolumn{2}{c}{Min-Max} & \multicolumn{2}{c}{WP} \\
\midrule
\multicolumn{5}{l}{\emph{FW2-Med}} \\
No rephrasing                                & 57.16 & ${}_{\pm 14.55}$ & 53.97 & ${}_{\pm 5.20}$ \\
\midrule
\multicolumn{5}{l}{\emph{Rephrased only}} \\
Qwen                                         & 63.47 & ${}_{\pm 6.65}$ & 57.14 & ${}_{\pm 7.53}$ \\
Gemma                                        & 54.03 & ${}_{\pm 6.41}$ & 47.62 & ${}_{\pm 8.25}$ \\
Qwen + Gemma (1:1)                         & 54.75 & ${}_{\pm 13.13}$ & 49.21 & ${}_{\pm 4.20}$ \\
Qwen, density-filtered                      & 44.17 & ${}_{\pm 11.16}$ & 33.33 & ${}_{\pm 7.53}$ \\
\midrule
\multicolumn{5}{l}{\emph{Rephrased + raw}} \\
Qwen + raw                                   & \underline{70.15} & ${}_{\pm 9.51}$ & \underline{65.08} & ${}_{\pm 5.38}$ \\
Qwen + filtered raw                          & \textbf{81.55} & ${}_{\pm 7.16}$ & \textbf{77.78} & ${}_{\pm 4.20}$ \\
\quad Qwen:filtered raw = 2:1                & 47.12 & ${}_{\pm 14.50}$ & 44.44 & ${}_{\pm 6.04}$ \\
\quad Qwen:filtered raw = 1:1                & 60.25 & ${}_{\pm 7.96}$ & 52.38 & ${}_{\pm 5.83}$ \\
\quad Qwen:filtered raw = 1:2                & 19.97 & ${}_{\pm 8.16}$ & 19.05 & ${}_{\pm 4.12}$ \\
\bottomrule
\end{tabular}
\caption{Mix-variant ablation at the 1M-source-document scale: the strongest configuration mixes Qwen-rephrased data with filtered raw data. Qwen and Gemma denote Qwen3.5-35B-A3B and Gemma-4-26B-A4B. \emph{Filtered raw} applies the Edu $\cap$ Med-term filter from §\ref{sec:exp-filtering} to the raw side. Indented rows downsample the rephrased side to vary the Qwen:filtered-raw ratio. Metrics and formatting as in Table~\ref{tab:filter-ablation}. Per-task scores in Table~\ref{tab:rephrase-1M-per-task}.}
\label{tab:rephrase-20b}
\end{table}

We ablate at two source-document scales.
At 100k (Table~\ref{tab:rephrase-100k}), we test our recipe across five LLMs and standard MGA (re-run with one (genre, audience) pair per document for corpus-size parity).
The 1M ablation (Table~\ref{tab:rephrase-20b}) validates the strongest 100k recipes at higher compute and tests mixing with the best filtered raw data.
The raw baseline (unfiltered, unrephrased FW2-Med) is included at both scales.

\paragraph{Signal-amplifying rephrasing beats both raw and standard MGA.}
At 100k, our best rephraser outperforms raw by +40 WP points, while standard MGA with the same LLM falls below raw by 29 WP points.
This gap shows the medical adaptations of \S\ref{sec:rephrasing}, not the LLM, drive the gain.
At 1M, the rephrasing-alone gain attenuates: only Qwen-rephrased stays above raw, while Gemma-rephrased (which improved over raw at 100k) falls below.

\paragraph{Rephraser quality is not predicted by model scale or medical tuning.}
At 100k, the smaller Qwen3.5-35B-A3B outperforms its larger Qwen sibling, consistent with~\citet{niklaus_how_2026}'s rephraser-scale observations.
Medically-tuned MedGemma-27B lands far below generic Gemma-4-26B-A4B at similar parameter count, and GPT-OSS-120B collapses.
At 1M, we rule out two alternatives for the final recipe: a 50/50 Qwen+Gemma mix aimed at rephraser diversity does not gain over pure Qwen, and filtering the rephrased corpus by medical-term density underperforms pure rephrased, likely because rephrasing already encodes the density signal.

\paragraph{Rephrasing complements raw; filtering raw amplifies the gain.}
At 1M, mixing Qwen-rephrased with raw outperforms either alone, and the best filter (educational quality and medical-term density) on the raw side increases the gain to +24 WP points over raw.
We tested reduced Qwen:filtered-raw ratios to find an optimal mix, but none beats the full mix.

Together, \S\ref{sec:exp-filtering} and \S\ref{sec:exp-rephrasing} establish that filtering and rephrasing are complementary mechanisms: filtering raises per-token entity density, rephrasing raises per-entity context diversity, and combining them beats either alone.

\section{FineMed and DoctoBERT}
\label{sec:final}

We apply the validated \S\ref{sec:exp} recipe at scale to release the French medical pretraining corpora \emph{FineMed} and \emph{FineMed-rephrased} (\S\ref{sec:final-dataset}), and \emph{DoctoBERT}, an encoder family pretrained on them (\S\ref{sec:final-doctobert}).

\subsection{FineMed and FineMed-rephrased}
\label{sec:final-dataset}

\begin{table}[t]
\centering
\footnotesize
\setlength{\tabcolsep}{3pt}
\begin{tabular}{lrrrrr}
\toprule
Corpus & Docs & Words & Length & Edu & Density \\
\midrule
\multicolumn{6}{l}{\emph{Curated medical corpora}} \\
NACHOS                        & 2.4M  & 1.3B  & 11   & 1.32 & 0.374 \\
TransCorpus-bio-fr            & 21.6M & 5.3B  & 243  & 4.10 & 0.199 \\
\midrule
\multicolumn{6}{l}{\emph{Ours}} \\
FineMed                       & 21.1M & 19.2B & 369  & 2.09 & 0.079 \\
FineMed-filtered              & 2.1M  & 3.8B  & 665  & 4.37 & 0.198 \\
FineMed-rephrased             & 13.6M & 4.5B  & 191  & 2.86 & 0.164 \\
\bottomrule
\end{tabular}
\caption{Corpus statistics for \emph{FineMed} variants and curated medical corpora. Length is the median word count per document; Edu (educational-quality score) and Density (medical-term density) are document means. Per-source breakdown in Appendix~\ref{app:per-source-stats}.}
\label{tab:dataset-stats}
\end{table}

In \S\ref{sec:exp} we ablated filtering and rephrasing on a single source (FW2-Med) at limited scale. We now apply the same recipe to all three medical-prefiltered corpora (FineWeb-2, FinePDFs, FineWiki) at full scale. The multi-source assembly is a scaling step over the validated recipe, not a separately ablated design choice.

\textit{FineMed.}
We scale the multi-axis annotation of \S\ref{sec:annotation} (subdomain, educational quality, medical-term density) across all French medical content from the three sources.
\emph{FineMed}, the resulting corpus, is an order of magnitude larger than curated medical corpora (Table~\ref{tab:dataset-stats}) and is released unfiltered alongside three annotators to support task-specific filtering.
For medical encoder pretraining, we apply the strongest filter from \S\ref{sec:exp-filtering} (educational quality and medical-term density) to obtain \emph{FineMed-filtered}.

\textit{FineMed-rephrased.}
We apply the signal-amplifying rephrasing recipe of \S\ref{sec:rephrasing} to \emph{FineMed} to produce \emph{FineMed-rephrased}, which more than doubles medical-term density.
We also add a coarse pre-screen that reuses the existing multi-axis annotations to bypass rephrasing stage~1 for documents likely to be discarded, cutting LLM cost (Appendix~\ref{app:rephrasing-proxy}).
\subsection{DoctoBERT}
\label{sec:final-doctobert}

We release two members: \emph{DoctoBERT-fr}, a classic RoBERTa encoder, and \emph{DoctoModernBERT-fr}, a more efficient, long-context ModernBERT encoder.

\textit{Tokenizer.}
We train a SentencePiece BPE tokenizer on the entity-rich \emph{FineMed-filtered} subset.
Vocabulary size matches each backbone's default (50k for ModernBERT, 32k for RoBERTa; Appendix~\ref{app:tokenizer}).

\textit{DoctoModernBERT-fr.}
Following the ModernBERT recipe~\citep{warner_smarter_2024}, we train across three phases for a total of 240B tokens (Appendix~\ref{app:pretraining}).
P1 pretrains at 1024-token context on the \emph{FineMed-filtered} and \emph{FineMed-rephrased} mix to produce the base contextual representations.
P2 extends the context window to 8192 tokens on a subset upsampled toward longer documents.
P3 anneals on the biomedical and clinical subdomains (Bio\&Cli; \S\ref{sec:subdomain}) of the mix to focus the final updates on content closest to downstream medical use.

\textit{DoctoBERT-fr.}
We use the RoBERTa architecture~\citep{liu_roberta_2019}, training in two phases: a 500B-token pretraining phase on the same mix, then a 200B-token annealing phase on the Bio\&Cli subset, as in \emph{DoctoModernBERT-fr}'s final phase.

\begin{table*}[!htbp]
\centering
\scriptsize
\setlength{\tabcolsep}{3pt}
\begin{tabular}{lrrrrrrr r@{}l r@{}l}
\toprule
 & \multicolumn{2}{c}{QUAERO} & \multicolumn{2}{c}{E3C} & MORFITT & DEFT2021 & DIAMED & \multicolumn{4}{c}{Aggregate} \\
\cmidrule(lr){2-3} \cmidrule(lr){4-5} \cmidrule(lr){6-6} \cmidrule(lr){7-7} \cmidrule(lr){8-8} \cmidrule(lr){9-12}
Model & EMEA & MEDLINE & CLIN. & TEMP. & CLS & NER & CLS & \multicolumn{2}{c}{Min-Max} & \multicolumn{2}{c}{WP} \\
\midrule
\multicolumn{12}{l}{\emph{English medical}} \\
BioBERT                & 58.77$_{\pm 1.50}$ & 50.29$_{\pm 0.61}$ & 55.02$_{\pm 1.63}$ & 78.29$_{\pm 0.75}$ & 66.99$_{\pm 0.97}$ & 56.72$_{\pm 0.60}$ & 59.26$_{\pm 1.29}$ & 29.97 & ${}_{\pm 6.94}$ & 15.71 & ${}_{\pm 9.63}$ \\
BioClinical-ModernBERT & 44.74$_{\pm 2.50}$ & 44.44$_{\pm 3.51}$ & 49.53$_{\pm 1.14}$ & 76.11$_{\pm 1.12}$ & 67.42$_{\pm 1.35}$ & 53.97$_{\pm 1.63}$ & 52.07$_{\pm 4.95}$ & 0.88 & ${}_{\pm 0.88}$ & 1.43 & ${}_{\pm 1.43}$ \\
ModernBERT-bio         & 56.84$_{\pm 1.67}$ & 46.60$_{\pm 0.44}$ & 53.76$_{\pm 0.86}$ & 78.85$_{\pm 0.49}$ & 68.57$_{\pm 0.95}$ & 56.43$_{\pm 0.94}$ & 61.06$_{\pm 1.50}$ & 29.35 & ${}_{\pm 5.86}$ & 17.14 & ${}_{\pm 10.17}$ \\
\midrule
\multicolumn{12}{l}{\emph{French generalist}} \\
CamemBERT              & 65.43$_{\pm 0.96}$ & 56.18$_{\pm 1.00}$ & 59.82$_{\pm 0.71}$ & 83.81$_{\pm 0.47}$ & 71.54$_{\pm 0.22}$ & 62.40$_{\pm 0.36}$ & 60.26$_{\pm 2.28}$ & 69.37 & ${}_{\pm 6.39}$ & 57.14 & ${}_{\pm 13.30}$ \\
ModernCamemBERT        & 61.98$_{\pm 1.27}$ & 55.46$_{\pm 1.05}$ & 57.62$_{\pm 0.81}$ & 83.11$_{\pm 0.37}$ & 70.01$_{\pm 0.97}$ & 60.01$_{\pm 1.24}$ & 53.26$_{\pm 2.12}$ & 52.69 & ${}_{\pm 9.19}$ & 28.57 & ${}_{\pm 13.64}$ \\
\midrule
\multicolumn{12}{l}{\emph{French medical}} \\
DrBERT                 & 64.37$_{\pm 1.05}$ & 57.18$_{\pm 0.48}$ & 58.01$_{\pm 0.79}$ & 82.44$_{\pm 1.05}$ & 70.42$_{\pm 0.45}$ & 61.08$_{\pm 0.74}$ & 64.87$_{\pm 1.90}$ & 65.08 & ${}_{\pm 4.15}$ & 44.29 & ${}_{\pm 14.51}$ \\
CamemBERT-bio          & 64.98$_{\pm 1.03}$ & 59.03$_{\pm 0.63}$ & 61.40$_{\pm 0.66}$ & \textbf{84.88$_{\pm 0.59}$} & 71.48$_{\pm 0.67}$ & 64.73$_{\pm 0.46}$ & 64.63$_{\pm 2.89}$ & 80.83 & ${}_{\pm 5.31}$ & 70.00 & ${}_{\pm 10.95}$ \\
TransBERT-bio-fr       & \underline{67.37$_{\pm 1.24}$} & \underline{59.96$_{\pm 0.35}$} & \underline{62.36$_{\pm 0.91}$} & 84.48$_{\pm 0.32}$ & \textbf{74.04$_{\pm 0.63}$} & \underline{65.48$_{\pm 0.53}$} & 70.91$_{\pm 2.23}$ & \underline{93.88} & ${}_{\pm 1.86}$ & \underline{88.57} & ${}_{\pm 8.46}$ \\
ModernCamemBERT-bio    & 65.35$_{\pm 0.45}$ & 56.81$_{\pm 0.99}$ & 58.63$_{\pm 0.50}$ & 83.31$_{\pm 0.57}$ & 71.21$_{\pm 0.37}$ & 61.35$_{\pm 0.55}$ & 67.77$_{\pm 3.44}$ & 71.37 & ${}_{\pm 4.11}$ & 54.29 & ${}_{\pm 14.72}$ \\
\midrule
\multicolumn{12}{l}{\emph{Ours}} \\
\textbf{DoctoBERT-fr}    & \textbf{68.39$_{\pm 0.84}$} & \textbf{62.54$_{\pm 0.45}$} & \textbf{62.75$_{\pm 1.62}$} & \underline{84.60$_{\pm 0.51}$} & \underline{73.36$_{\pm 0.26}$} & \textbf{66.41$_{\pm 0.43}$} & \textbf{72.56$_{\pm 1.23}$} & \textbf{98.17} & ${}_{\pm 1.38}$ & \textbf{97.14} & ${}_{\pm 1.90}$ \\
\textbf{DoctoModernBERT-fr} & 65.71$_{\pm 0.51}$ & 59.65$_{\pm 0.40}$ & 59.62$_{\pm 0.57}$ & 84.06$_{\pm 0.62}$ & 71.87$_{\pm 0.92}$ & 63.81$_{\pm 0.63}$ & \underline{71.60$_{\pm 4.14}$} & 83.15 & ${}_{\pm 3.43}$ & 75.71 & ${}_{\pm 12.24}$ \\
\bottomrule
\end{tabular}
\caption{Per-task and aggregate DrBenchmark results. \emph{DoctoBERT-fr} wins both aggregate metrics and five of seven per-task scores. Per-task cells are mean$\pm$std F1 on the test split; aggregate cells are mean$\pm$SE across tasks. Best per column in bold, second underlined. Tasks: 5 NER (QUAERO EMEA/MEDLINE, E3C Clinical/Temporal, DEFT2021) and 2 classification (MORFITT, DIAMED). Metrics as in Table~\ref{tab:filter-ablation}.}
\label{tab:final}
\end{table*}

\begin{table}[!htbp]
\centering
\scriptsize
\setlength{\tabcolsep}{3pt}
\begin{tabular}{lrrr}
\toprule
Model & Precision & Recall & F1 \\
\midrule
\multicolumn{4}{l}{\emph{English medical}} \\
BioBERT                & 77.54$_{\pm 0.78}$ & 78.42$_{\pm 1.01}$ & 77.97$_{\pm 0.29}$ \\
BioClinical-ModernBERT & \underline{78.79$_{\pm 0.42}$} & 78.69$_{\pm 0.70}$ & 78.74$_{\pm 0.54}$ \\
ModernBERT-bio         & 78.06$_{\pm 1.40}$ & 79.30$_{\pm 0.48}$ & 78.67$_{\pm 0.48}$ \\
\midrule
\multicolumn{4}{l}{\emph{French generalist}} \\
CamemBERT              & 77.19$_{\pm 0.87}$ & 79.58$_{\pm 1.64}$ & 78.36$_{\pm 0.37}$ \\
ModernCamemBERT        & 78.53$_{\pm 0.14}$ & 78.71$_{\pm 0.73}$ & 78.62$_{\pm 0.43}$ \\
\midrule
\multicolumn{4}{l}{\emph{French medical}} \\
DrBERT                 & 76.77$_{\pm 0.87}$ & 77.81$_{\pm 0.87}$ & 77.28$_{\pm 0.36}$ \\
CamemBERT-bio          & 77.51$_{\pm 0.84}$ & 78.90$_{\pm 1.00}$ & 78.19$_{\pm 0.19}$ \\
TransBERT-bio-fr       & 76.85$_{\pm 1.06}$ & 78.66$_{\pm 0.92}$ & 77.74$_{\pm 0.25}$ \\
ModernCamemBERT-bio    & 78.17$_{\pm 0.46}$ & \textbf{79.76$_{\pm 1.19}$} & \underline{78.95$_{\pm 0.41}$} \\
\midrule
\multicolumn{4}{l}{\emph{Ours}} \\
\textbf{DoctoBERT-fr}    & 77.29$_{\pm 0.16}$ & 79.68$_{\pm 0.63}$ & 78.47$_{\pm 0.27}$ \\
\textbf{DoctoModernBERT-fr} & \textbf{79.12$_{\pm 1.42}$} & \underline{79.71$_{\pm 1.04}$} & \textbf{79.40$_{\pm 0.22}$} \\
\bottomrule
\end{tabular}
\caption{Performance on the proprietary clinical NER task: mean$\pm$std over 3 seeds on the held-out test split. Best in bold, second underlined.}
\label{tab:realworld-ner}
\end{table}

\subsection{Evaluation}
\label{sec:final-eval}

We evaluate on two complementary tasks: DrBenchmark~\citep{labrak_drbenchmark_2024}, the public French biomedical benchmark, and a proprietary French clinical NER task from a real-world production setting.

\textit{Baselines.}
We compare against nine encoders in three groups: four French medical encoders~\citep{knafou_transbert_2025, labrak_drbert_2023, touchent_camembertbio_2024, touchent_causal_2026}; two French generalist encoders~\citep{antoun_modernbert_2025, martin_camembert_2020} to test whether medical-domain pretraining adds value over French-language pretraining alone; and three English medical encoders~\citep{lee_biobert_2020, sounack_bioclinical_2025, touchent_causal_2026} to test cross-lingual transfer from English medical pretraining.

\textit{DrBenchmark.}
On the public benchmark, \emph{DoctoBERT-fr} leads both overall aggregate metrics (Table~\ref{tab:final}).
Per task, \emph{DoctoBERT-fr} leads on 4 of 5 NER tasks (QUAERO EMEA/MEDLINE, E3C Clinical, DEFT2021) and on DIAMED classification. TransBERT-bio-fr retains MORFITT (biomedical specialty classification, matching its PubMed-based pretraining).

\textit{Real-world clinical NER.}
We additionally evaluate on a proprietary French clinical NER task from a real-world production setting, beyond academic benchmarks.
The task spans varied clinical text, from consultation summaries to short structured patient-record entries.
The taxonomy covers 12 entity classes (e.g., pathology, drug, exam, biometry) and 9 qualifier classes (e.g., negation, family relationship, date; Appendix~\ref{app:real-world-ner}).
\emph{DoctoModernBERT-fr} achieves the highest precision and F1 (Table~\ref{tab:realworld-ner}), supporting the heterogeneous-web design: broader register coverage than curated-corpus baselines translates to better performance in a real-world production setting.
\section{Conclusion}
\label{sec:conclusion}

We show that decoder LLM web-data curation methodology can be adapted for encoder MLM pretraining in dense-terminology domains.
Through automated annotation and rephrasing, our recipe builds a large French medical corpus from heterogeneous web data, without manual curation or English translation.
The resulting \emph{DoctoBERT} encoders advance the state of the art on DrBenchmark and a proprietary real-world clinical NER task.

\section*{Limitations}
\label{sec:limitations}

\textit{Language and domain scope.}
We instantiate the recipe on French medical NLP.
The pipeline is language- and domain-agnostic by design (multilingual LLM teachers and rephrasers).
The recipe instances (subdomain taxonomy, educational-quality scoring rubric, medical entity classes, LLM rephrasing prompts) are medical-by-construction, and the distilled small annotators we ship are French-specific.
For a new language, re-distilling the annotators suffices; a new terminology-dense domain additionally requires redesigning these instances.
The underlying signals (per-token entity density, per-entity context diversity) are not inherently medical, though we do not test cross-domain transfer empirically.

\textit{Architecture and training objective.}
This work focuses on data curation rather than model design.
The recipe is selected via ModernBERT ablations and applied at full scale to two MLM encoder backbones at around 110--150M parameters, ModernBERT~\citep{warner_smarter_2024} and RoBERTa~\citep{liu_roberta_2019}.
Extensions to alternative architectures (e.g., NeoBERT~\citep{breton_neobert_2025}, EuroBERT~\citep{boizard_eurobert_2025}), alternative training objectives (biphasic Causal Language Modeling (CLM) then MLM~\citep{gisserot-boukhlef_should_2025, touchent_causal_2026}, decoder-to-encoder conversion~\citep{openai2025privacyfilter}), other parameter scales, and decoder LLM pretraining are open follow-ups.

\textit{Long-context evaluation.}
\emph{DoctoModernBERT-fr} inherits ModernBERT's 8192-token context window, but the current downstream evaluation only consists of short-context tasks and cannot evaluate long-context performance.
Building a long-context French medical benchmark would address this gap.

\section*{Ethical considerations}
\label{sec:ethics}

\textit{Data sources and PII/PHI.}
We pretrain on three public web corpora under permissive licenses (FineWeb-2 and FinePDFs under ODC-By 1.0; FineWiki adapted from Wikipedia under CC BY-SA 4.0). These sources may contain PII and, in medical pages, Protected Health Information (PHI), all already publicly accessible. We do not add de-identification on the raw side. The rephrasing pipeline instructs the LLM to replace non-medical PII with varied fictional values and to preserve medical content faithfully (\S\ref{sec:rephrasing}). Downstream consumers handling PHI should apply task-appropriate anonymization.

\section*{Acknowledgments}
We thank Yanis Labrak and Adrien Bazoge for helpful discussions in the early stages of this work, Adrien Giraud for valuable discussions, and the MFE team at Doctolib (Issa Ka, Foucauld Estignard, Cyril Laitang, and Nicolas Perquier) for enabling the evaluation on a real-world clinical task beyond academic benchmarks. This work was granted access to the HPC resources of IDRIS under the allocations 2025-AD011016291 and 2026-A0200617487 made by GENCI.

\bibliography{references,references_non_zotero}

\clearpage
\appendix

\section{Medical-Content Prefiltering}
\label{app:source-data}

\paragraph{Classifier and inference.}
We use a pretrained multilingual domain classifier~\citep{nvidia_multilingual_domain_classifier}, a fine-tuned DeBERTa-v3 covering 26 domains.
We apply it to the first 512 tokens of each document due to the model's context length, and retain documents whose top-1 predicted domain is \textsc{Health}.
To minimize padding overhead at scale, we group examples by length into buckets and sort within each bucket before batch inference.

\begin{table*}[t]
\centering
\footnotesize
\setlength{\tabcolsep}{6pt}
\begin{tabular}{lccc}
\toprule
Corpus (FR) & Raw (docs / words) & Medical (docs / words) & Frac.\ (docs / words) \\
\midrule
FineWeb-2 & 360M / 189.7B & 18.9M / 12.0B & 5.3\% / 6.3\% \\
FinePDFs  & 27.3M / 88.3B & 2.1M / 7.2B   & 7.7\% / 8.1\% \\
FineWiki  & 2.57M / 1.86B & 38.6k / 26.6M & 1.5\% / 1.4\% \\
\bottomrule
\end{tabular}
\caption{Per-source medical-content retention on the French split. The \emph{Frac.\ (docs / words)} column reports the share retained by document and by word count.}
\label{tab:prefilter-funnel}
\end{table*}

\paragraph{Per-domain distribution.}
Medical content is a small fraction of each source (Table~\ref{tab:domain-distribution}). FineWeb-2 is dominated by general-interest domains (Arts \& Entertainment, Home \& Garden, News, People \& Society); FinePDFs by Law \& Government and Jobs \& Education (public-administration PDFs); FineWiki by Arts \& Entertainment, People \& Society, Sports, and Travel \& Transportation.

\begin{table}[t]
\centering
\scriptsize
\setlength{\tabcolsep}{3pt}
\begin{tabular}{lrrr}
\toprule
Domain & FineWeb-2 & FinePDFs & FineWiki \\
\midrule
Adult                       & 8.8M  (2.5\%)  & 25.8k (0.1\%)  & 4.9k   (0.2\%)  \\
Arts \& Entertainment       & 35.0M (9.7\%)  & 1.76M (6.5\%)  & 478.4k (18.6\%) \\
Autos \& Vehicles           & 8.9M  (2.5\%)  & 405k  (1.5\%)  & 35.1k  (1.4\%)  \\
Beauty \& Fitness           & 13.0M (3.6\%)  & 219k  (0.8\%)  & 5.9k   (0.2\%)  \\
Books \& Literature         & 13.9M (3.9\%)  & 833.8k (3.1\%) & 101.4k (3.9\%)  \\
Business \& Industrial      & 18.1M (5.0\%)  & 1.56M (5.7\%)  & 23.7k  (0.9\%)  \\
Computers \& Electronics    & 13.3M (3.7\%)  & 662.9k (2.4\%) & 26.6k  (1.0\%)  \\
Finance                     & 7.3M  (2.0\%)  & 575.2k (2.1\%) & 8.5k   (0.3\%)  \\
Food \& Drink               & 20.2M (5.6\%)  & 1.35M (5.0\%)  & 43.3k  (1.7\%)  \\
Games                       & 11.7M (3.2\%)  & 181k  (0.7\%)  & 32.8k  (1.3\%)  \\
\textbf{Health}             & \textbf{18.9M (5.2\%)} & \textbf{2.14M (7.8\%)} & \textbf{38.6k (1.5\%)} \\
Hobbies \& Leisure          & 14.8M (4.1\%)  & 736.9k (2.7\%) & 22.2k  (0.9\%)  \\
Home \& Garden              & 23.6M (6.5\%)  & 646k  (2.4\%)  & 4.9k   (0.2\%)  \\
Internet \& Telecom         & 7.6M  (2.1\%)  & 193.6k (0.7\%) & 8.6k   (0.3\%)  \\
Jobs \& Education           & 14.8M (4.1\%)  & 3.19M (11.7\%) & 19.4k  (0.8\%)  \\
Law \& Government           & 8.4M  (2.3\%)  & 4.51M (16.5\%) & 86.1k  (3.4\%)  \\
News                        & 22.0M (6.1\%)  & 1.17M (4.3\%)  & 148.4k (5.8\%)  \\
Online Communities          & 5.0M  (1.4\%)  & 20.4k (0.1\%)  & 10.1k  (0.4\%)  \\
People \& Society           & 21.2M (5.9\%)  & 2.77M (10.2\%) & 466.1k (18.2\%) \\
Pets \& Animals             & 5.2M  (1.5\%)  & 269.7k (1.0\%) & 116.9k (4.6\%)  \\
Real Estate                 & 5.6M  (1.6\%)  & 409.4k (1.5\%) & 16.8k  (0.7\%)  \\
Science                     & 3.9M  (1.1\%)  & 1.08M (3.9\%)  & 186.4k (7.3\%)  \\
Sensitive Subjects          & 10.7M (3.0\%)  & 382.0k (1.4\%) & 49.9k  (1.9\%)  \\
Shopping                    & 9.8M  (2.7\%)  & 107.4k (0.4\%) & 3.9k   (0.2\%)  \\
Sports                      & 19.7M (5.5\%)  & 1.30M (4.8\%)  & 333.4k (13.0\%) \\
Travel \& Transportation    & 18.6M (5.2\%)  & 816.4k (3.0\%) & 293.8k (11.5\%) \\
\bottomrule
\end{tabular}
\caption{Document counts per domain on each source's French split. Cells show \emph{count (\%~of corpus)}. Health is one of 26 domains and a minority in every source.}
\label{tab:domain-distribution}
\end{table}

\section{Multi-Axis Annotation}
\label{app:filtering-classifiers}

\subsection{Subdomain Classifier}
\label{app:subdomain-classifier}

\paragraph{Taxonomy.}
We build the 15-class taxonomy through three rounds of LLM-driven iteration (full taxonomy in Table~\ref{tab:subdomain-taxonomy}).
Starting from a manually-defined initial taxonomy, we prompt Qwen3-235B-A22B-Instruct~\citep{yang_qwen3_2025} to annotate 10k random documents with reasoning traces.
We examine the per-class distribution and reasoning traces, merge or split poorly separated classes, and revise the prompt (final prompt in Prompt~\ref{app:subdomain-prompt}).

\begin{table*}[t]
\centering
\footnotesize
\setlength{\tabcolsep}{4pt}
\renewcommand{\arraystretch}{1.15}
\begin{tabular}{lp{10.5cm}}
\toprule
Class & Description \\
\midrule
Clinical cases \& vignettes              & Single-patient narratives covering presentation, evaluation, management, and outcomes; case-based teaching. \\
Clinical guidelines \& pathways          & Non-patient-specific recommendations, algorithms, and standards for evaluation, diagnosis, procedures, and treatment; named guidelines or consensus statements. \\
Patient education \& lifestyle           & Consumer-facing explanations and how-to advice on prevention, self-care, symptoms, diet, fitness, and mental well-being. \\
Wellness, supplements \& CAM             & Botanicals, vitamins, supplements, complementary or alternative therapies; claims outside mainstream clinical guidance and drug regulation. \\
Public health, policy \& programs        & Population surveillance, epidemiology, screening programs, laws/regulation, financing and insurance, community guidance. \\
Commercial \& promotional                & Marketing or sales content for products, services, or facilities: pricing, booking, calls-to-action, affiliate/SEO, comparative ads, testimonials. \\
Drugs, trials \& regulation              & Drug development and evaluation: clinical trials, approvals and labels, PK/PD, safety monitoring, pharmacovigilance. \\
Biomedical \& mechanistic science        & Experimental or preclinical research: labs, omics, pathways, cell/animal models, assays, mechanistic explanations. \\
Medical devices, diagnostics \& imaging  & Device or modality descriptions and clinical use; diagnostics, wearables, sensors, imaging. \\
Health IT, telemedicine \& operations    & EHR/EMR, data standards, interoperability, analytics, telemedicine platforms, workflow design, staffing, procurement, and supply-chain logistics. \\
Occupational health \& safety            & Workplace hazards, exposures, PPE, training, and compliance with occupational regulations. \\
Health workforce education \& training   & Professional curricula, CME, certification, simulation, residency or fellowship information. \\
Health services \& facilities            & Neutral descriptions of care delivery models, service lines, facility capabilities, and long-term or residential care. \\
Other health                             & Health-related content unclear or insufficient to classify under other topics. \\
Others                                   & Content that is not clearly health-related, is too brief, or lacks sufficient detail to classify under the other topics. \\
\bottomrule
\end{tabular}
\caption{Medical subdomain taxonomy (15 classes) used in LLM annotation and classifier finetuning.}
\label{tab:subdomain-taxonomy}
\end{table*}

\paragraph{LLM annotation.}
We collect supervision in two stages: Qwen3-30B-A3B-Instruct annotates 1M documents (stage-1); Qwen3-235B-A22B-Instruct annotates 500k, with 490k for stage-2 fine-tuning and 10k held out for evaluation.
Annotation uses content and URL as input, with shuffled class order to mitigate position bias.

\paragraph{Schedule ablation.}
We test two variants of the schedule: a smaller stage-2 (90k), and stage-2-only (no stage-1).
Expanding stage-2 supervision from 90k to 490k annotations marginally lifts macro F1 (Table~\ref{tab:subdomain-stage-f1}), with the largest gains on under-represented classes (clinical guidelines, drugs, biomedical).
A stage-2-only model (no stage-1 initialization) reaches similar overall F1.
We keep the two-stage variant because the held-out set is also annotated with Qwen3-235B-A22B-Instruct, and a stage-2-only model would be biased toward this teacher.

\paragraph{Training and evaluation.}
We fine-tune ModernCamemBERT-base on stage-1, then on stage-2 supervision (8192-token input).
Per-class precision, recall, and F1 are in Table~\ref{tab:subdomain-perclass}.

\begin{table}[t]
\centering
\footnotesize
\setlength{\tabcolsep}{4pt}
\begin{tabular}{lccc}
\toprule
Stage & Prec.\ & Recall & F1 \\
\midrule
Stage 1 (1M docs)              & 0.71 & 0.70 & 0.69 \\
Stage 1 + Stage 2 (90k docs)   & 0.74 & 0.74 & 0.74 \\
Stage 1 + Stage 2 (490k docs)  & 0.75 & 0.75 & 0.75 \\
Stage 2 only (490k docs)       & 0.74 & 0.75 & 0.74 \\
\bottomrule
\end{tabular}
\caption{Subdomain classifier weighted-average results on the 10k held-out evaluation set across training stages. Stage 1 uses the Qwen3-30B-A3B-Instruct teacher; Stage 2 uses the Qwen3-235B-A22B-Instruct teacher. Stage 2 lifts F1.}
\label{tab:subdomain-stage-f1}
\end{table}

\begin{table}[t]
\centering
\scriptsize
\setlength{\tabcolsep}{3pt}
\begin{tabular}{lrrrr}
\toprule
Class & Prec.\ & Rec.\ & F1 & Sup.\ \\
\midrule
Biomedical \& mechanistic science       & 0.72 & 0.69 & 0.70 & 296   \\
Clinical cases \& vignettes             & 0.65 & 0.62 & 0.63 & 234   \\
Clinical guidelines \& pathways         & 0.70 & 0.54 & 0.61 & 379   \\
Commercial \& promotional               & 0.76 & 0.74 & 0.75 & 1{,}416 \\
Drugs, trials \& regulation             & 0.66 & 0.66 & 0.66 & 298   \\
Health IT, telemedicine \& operations   & 0.66 & 0.74 & 0.70 & 193   \\
Health services \& facilities           & 0.67 & 0.68 & 0.68 & 711   \\
Health workforce education \& training  & 0.75 & 0.75 & 0.75 & 507   \\
Medical devices, diagnostics \& imaging & 0.72 & 0.73 & 0.73 & 478   \\
Occupational health \& safety           & 0.75 & 0.79 & 0.77 & 313   \\
Other health                            & 0.00 & 0.00 & 0.00 & 22    \\
Others                                  & 0.58 & 0.59 & 0.58 & 404   \\
Patient education \& lifestyle          & 0.78 & 0.80 & 0.79 & 1{,}984 \\
Public health, policy \& programs       & 0.84 & 0.86 & 0.85 & 2{,}036 \\
Wellness, supplements \& CAM            & 0.71 & 0.73 & 0.72 & 729   \\
\midrule
Weighted average                        & 0.75 & 0.75 & 0.75 & 10{,}000 \\
\bottomrule
\end{tabular}
\caption{Per-class subdomain classifier results on the 10k held-out set.}
\label{tab:subdomain-perclass}
\end{table}

\paragraph{Per-corpus distribution.}
Subdomain distributions differ markedly across the three corpora (Table~\ref{tab:subdomain-distribution}).
FineWeb-2 leans toward consumer-facing content (\emph{Patient education \& lifestyle}, \emph{Commercial \& promotional}, \emph{Wellness, supplements \& CAM}); FinePDFs toward institutional content (\emph{Public health, policy \& programs}, \emph{Clinical guidelines \& pathways}); FineWiki toward encyclopedic content (\emph{Biomedical \& mechanistic science}).

\begin{table}[t]
\centering
\scriptsize
\setlength{\tabcolsep}{3pt}
\begin{tabular}{lrrr}
\toprule
Subdomain & FineWeb-2 & FinePDFs & FineWiki \\
\midrule
Biomedical \& mechanistic science       & 2.7\%  & 2.8\%  & 23.9\% \\
Clinical cases \& vignettes             & 2.2\%  & 2.2\%  & 1.9\%  \\
Clinical guidelines \& pathways         & 3.3\%  & 9.1\%  & 13.0\% \\
Commercial \& promotional               & 14.5\% & 3.1\%  & 0.1\%  \\
Drugs, trials \& regulation             & 3.0\%  & 3.2\%  & 8.4\%  \\
Health IT, telemedicine \& operations   & 1.9\%  & 2.0\%  & 0.4\%  \\
Health services \& facilities           & 7.0\%  & 8.7\%  & 18.0\% \\
Health workforce education \& training  & 5.0\%  & 13.4\% & 6.7\%  \\
Medical devices, diagnostics \& imaging & 4.3\%  & 4.6\%  & 3.0\%  \\
Occupational health \& safety           & 3.4\%  & 8.1\%  & 0.8\%  \\
Patient education \& lifestyle          & 19.9\% & 8.7\%  & 3.6\%  \\
Public health, policy \& programs       & 21.9\% & 29.9\% & 13.9\% \\
Wellness, supplements \& CAM            & 7.4\%  & 1.6\%  & 1.6\%  \\
Others                                  & 3.4\%  & 2.5\%  & 4.5\%  \\
Other health                            & 0.0\%  & 0.0\%  & 0.2\%  \\
\bottomrule
\end{tabular}
\caption{Predicted subdomain distribution (\%) per corpus on the prefiltered medical subset.}
\label{tab:subdomain-distribution}
\end{table}

\subsection{Educational-Quality Scorer}
\label{app:edu-classifier}

\paragraph{Scoring rules.}
We adapt FineWeb-Edu's additive 0--5 scoring from a general school-education target to a medical-education target (medical students, residents, practicing clinicians, or other health professionals).
Each point is awarded for a successive criterion: minimal medical informativeness, usable information with low noise, domain specificity and density, structural coherence, and expert synthesis or actionable guidance.
We refine the scoring rules on 5k LLM-annotated documents based on score distribution and human spot-checks (final form in Prompt~\ref{app:edu-prompt}).

\paragraph{LLM annotation.}
Stage 1 uses Qwen3-30B-A3B-Instruct to annotate 1M documents; stage 2 uses Qwen3-235B-A22B-Instruct on 100k, with 90k for fine-tuning and 10k held out for evaluation.

\paragraph{Head and rounding ablation.}
We test two head choices (regression vs.\ classification) and three rounding strategies for the regression head (round-up, floor, ceil).
Regression outperforms classification on stage-1.
Round-up gives the best stage-2 F1 (Table~\ref{tab:edu-ablation}).

\paragraph{Training and evaluation.}
We fine-tune ModernCamemBERT-base in two stages at 8192-token input.
Unlike FineWeb-Edu, we do not freeze embeddings or the encoder.
Stage 2 lifts F1 over stage 1 alone (Table~\ref{tab:edu-stage-f1}).
Table~\ref{tab:edu-perclass} reports per-class metrics.

\begin{table}[t]
\centering
\footnotesize
\setlength{\tabcolsep}{4pt}
\begin{tabular}{lccc}
\toprule
Stage & Prec.\ & Recall & F1 \\
\midrule
Stage 1 (1M docs)              & 0.60 & 0.61 & 0.60 \\
Stage 1 + Stage 2 (90k docs)   & 0.67 & 0.66 & 0.66 \\
\bottomrule
\end{tabular}
\caption{Training-stage breakdown for the educational-quality scorer (regression head, round-up rounding), weighted averages on the 10k held-out set; Stage 2 lifts F1 over Stage 1 alone.}
\label{tab:edu-stage-f1}
\end{table}

\begin{table}[t]
\centering
\footnotesize
\setlength{\tabcolsep}{3pt}
\begin{tabular}{lccc}
\toprule
Configuration & Prec.\ & Recall & F1 \\
\midrule
\emph{Output head (Stage 1):} & & & \\
\quad Regression (round-up) & 0.60 & 0.61 & 0.60 \\
\quad Classification        & 0.58 & 0.59 & 0.56 \\
\midrule
\emph{Rounding (Stage 1+2):} & & & \\
\quad Round-up         & 0.67 & 0.66 & 0.66 \\
\quad Floor            & 0.53 & 0.48 & 0.47 \\
\quad Ceil             & 0.54 & 0.46 & 0.45 \\
\bottomrule
\end{tabular}
\caption{Output-head and rounding-strategy ablations on the 10k held-out set, weighted averages. Regression outperforms classification at Stage 1; among regression-head rounding strategies after Stage 1+2, round-up gives the best F1.}
\label{tab:edu-ablation}
\end{table}

\begin{table}[t]
\centering
\footnotesize
\setlength{\tabcolsep}{4pt}
\begin{tabular}{lrrrr}
\toprule
Score & Prec.\ & Rec.\ & F1 & Sup.\ \\
\midrule
0 & 0.73 & 0.54 & 0.62 & 902     \\
1 & 0.75 & 0.80 & 0.77 & 3{,}515 \\
2 & 0.64 & 0.62 & 0.63 & 2{,}554 \\
3 & 0.49 & 0.58 & 0.53 & 1{,}334 \\
4 & 0.59 & 0.58 & 0.59 & 1{,}152 \\
5 & 0.75 & 0.57 & 0.65 & 543     \\
\midrule
Weighted average  & 0.67 & 0.66 & 0.66 & 10{,}000 \\
\bottomrule
\end{tabular}
\caption{Per-score-class results for the educational-quality scorer (10k held-out set).}
\label{tab:edu-perclass}
\end{table}

\paragraph{Per-corpus distribution.}
Corpora differ in educational-quality distribution (Table~\ref{tab:edu-distribution}).
FineWeb-2 spreads across the score range while FinePDFs and FineWiki concentrate at the high end (scores 4--5).
The medical-prefiltered FineWeb-2 subset retains over 50\% of words at score $\geq$ 3 and over 70\% at score $\geq$ 2, well above FineWeb-Edu's 8\% and 36\% on general web.
Prefiltering already lifts quality.

\begin{table}[t]
\centering
\footnotesize
\setlength{\tabcolsep}{3pt}
\begin{tabular}{lrrr}
\toprule
Score & FineWeb-2 & FinePDFs & FineWiki \\
\midrule
0 & 6.1\%  & 2.9\%  & 1.1\%  \\
1 & 37.6\% & 24.2\% & 18.5\% \\
2 & 25.2\% & 20.7\% & 15.2\% \\
3 & 15.5\% & 16.2\% & 18.0\% \\
4 & 11.3\% & 18.7\% & 26.7\% \\
5 & 4.3\%  & 17.3\% & 20.5\% \\
\bottomrule
\end{tabular}
\caption{Educational-quality score distribution (\%) per corpus on the prefiltered medical subset.}
\label{tab:edu-distribution}
\end{table}

\subsection{Medical-Term-Density Extractor}
\label{app:medterm-extractor}

\paragraph{Character-level rationale.}
Word counts hide subword fragmentation: rare medical terms split into several subword tokens, and these fragments are what MLM masks.
Token-level density would instead tie the measure to a specific tokenizer, while characters stay tokenizer-independent.
Table~\ref{tab:char-token-example} shows three medical terms with the same word count but different token counts.

\begin{table}[t]
\centering
\footnotesize
\begin{tabular}{l r r}
\toprule
Term & Words & Tokens \\
\midrule
sang & 1 & 1 \\
pneumonectomie & 1 & 2 \\
œsophagogastroduodénoscopie & 1 & 5 \\
\bottomrule
\end{tabular}
\caption{Word vs.\ token counts for three medical terms (DoctoBERT tokenizer). Longer, rarer terms fragment into more subword tokens, which word-level density cannot see.}
\label{tab:char-token-example}
\end{table}

\paragraph{Taxonomy.}
We adapt UMLS (Unified Medical Language System) entity groups into an 8-class taxonomy (full taxonomy in Table~\ref{tab:medterm-taxonomy}).
We keep medical-term-rich classes and refine names and descriptions to reduce non-medical extractions.

\begin{table*}[t]
\centering
\footnotesize
\setlength{\tabcolsep}{4pt}
\renewcommand{\arraystretch}{1.15}
\begin{tabular}{lp{12cm}}
\toprule
Class & Description \\
\midrule
\texttt{disease}           & Pathological condition: disease, syndrome, infection, cancer, injury, symptom, clinical finding, mental disorder. \\
\texttt{drug}              & Chemical substance for therapy: prescription medication, vaccine, therapeutic compound, drug class, contrast agent. \\
\texttt{body\_part}        & Anatomical structure: organ, tissue, bone, muscle, blood vessel, nerve, cell, body fluid, anatomical region. \\
\texttt{medical\_procedure}& Clinical action with methodology: surgery, diagnostic test, medical examination, laboratory test, imaging procedure. \\
\texttt{molecular\_marker} & Molecular entity or biochemical substance: gene, protein, enzyme, receptor, genetic variant, biochemical analyte. \\
\texttt{clinical\_device}  & Manufactured medical object: surgical tool, implant, prosthetic, diagnostic scanner, monitoring equipment. \\
\texttt{vital\_function}   & Physiological parameter name: heart rate, blood pressure, respiratory rate, temperature, oxygen saturation. \\
\texttt{living\_beings}    & Non-human organism in biomedical context: bacterium, virus, fungus, parasite, pathogen, model organism. \\
\bottomrule
\end{tabular}
\caption{Medical entity taxonomy (8 classes) used in both Qwen3-235B-A22B-Instruct annotation and GLiNER2 finetuning. Adapted from UMLS by keeping medical-term-rich groups and tightening descriptions.}
\label{tab:medterm-taxonomy}
\end{table*}

\paragraph{LLM annotation.}
Single-pass entity extraction from teacher LLMs is unreliable, so Qwen3-235B-A22B-Instruct annotates roughly 300k documents via two-pass self-review: Pass 1 extracts entities under the 8-class taxonomy (Prompt~\ref{app:medterm-prompt-pass1}); Pass 2 reviews and corrects Pass 1's output (Prompt~\ref{app:medterm-prompt-pass2}).
We hold out 10k for evaluation, and both passes shuffle entity-group order to mitigate position bias.

\paragraph{Training-data and description ablation.}
We ablate train-time and inference-time conditions for the extractor: training size (pretrained, 100k, 300k samples), training prompts (with vs.\ without descriptions), and inference prompts (with vs.\ without descriptions).
Fine-tuning lifts F1 over pretrained GLiNER2.
Descriptions help most at inference but hurt when added to training prompts.
The best configuration: 300k training samples, training prompts without descriptions, inference prompts with descriptions (Table~\ref{tab:medterm-ablation}).

\paragraph{Training and evaluation.}
We fine-tune GLiNER2 on annotations reviewed by Qwen3-235B-A22B-Instruct, using the best ablation configuration.
The backbone is mDeBERTa-v3 with a 512-token context.
At inference, we truncate each document to its middle 512 tokens (rather than chunking). This skips boilerplate at document boundaries and keeps corpus-level inference tractable.
Per-class results are in Table~\ref{tab:medterm-perclass}.

\begin{table}[t]
\centering
\footnotesize
\setlength{\tabcolsep}{4pt}
\begin{tabular}{llccc}
\toprule
Train & Infer & Prec.\ & Recall & F1 \\
\midrule
Pretrained          & w/o desc & 0.51 & 0.43 & 0.45 \\
Pretrained          & w/ desc  & 0.53 & 0.45 & \textbf{0.48} \\
\midrule
100k, w/o desc      & w/o desc & 0.50 & 0.72 & 0.58 \\
100k, w/o desc      & w/ desc  & 0.62 & 0.73 & \textbf{0.67} \\
\midrule
100k, w/ desc       & w/o desc & 0.49 & 0.72 & 0.57 \\
100k, w/ desc       & w/ desc  & 0.48 & 0.71 & 0.57 \\
\midrule
300k, w/o desc      & w/o desc & 0.49 & 0.72 & 0.57 \\
300k, w/o desc      & w/ desc  & 0.66 & 0.73 & \textbf{0.69} \\
\bottomrule
\end{tabular}
\caption{GLiNER2 medical-entity extractor: F1 on the 10k held-out set across train-time supervision (pretrained, 100k samples, 300k samples; with or without per-class descriptions in the training prompts) and inference-time conditions (with or without descriptions). Descriptions help most at inference. Training without descriptions and inferring with them is the best configuration.}
\label{tab:medterm-ablation}
\end{table}

\begin{table}[t]
\centering
\footnotesize
\setlength{\tabcolsep}{4pt}
\begin{tabular}{lrrrr}
\toprule
Class & Prec.\ & Rec.\ & F1 & Sup.\ \\
\midrule
\texttt{body\_part}         & 0.70 & 0.82 & 0.76 & 12{,}454 \\
\texttt{clinical\_device}   & 0.52 & 0.70 & 0.60 & 4{,}252 \\
\texttt{disease}            & 0.71 & 0.77 & 0.74 & 21{,}678 \\
\texttt{drug}               & 0.65 & 0.76 & 0.70 & 7{,}794 \\
\texttt{living\_beings}     & 0.60 & 0.78 & 0.68 & 2{,}348 \\
\texttt{medical\_procedure} & 0.59 & 0.66 & 0.62 & 10{,}556 \\
\texttt{molecular\_marker}  & 0.68 & 0.64 & 0.66 & 8{,}092 \\
\texttt{vital\_function}    & 0.56 & 0.53 & 0.55 & 2{,}581 \\
\midrule
Weighted average            & 0.66 & 0.73 & 0.69 & 69{,}755 \\
\bottomrule
\end{tabular}
\caption{Per-class medical-entity extractor results on the 10k held-out set.}
\label{tab:medterm-perclass}
\end{table}

\paragraph{Per-corpus distribution.}
Medical-term density is low for most documents across all three corpora (Table~\ref{tab:density-distribution}).
FineWeb-2 and FinePDFs keep over 90\% of documents below 0.20 density; FineWiki is denser, with 16.0\% above 0.30.

\begin{table}[t]
\centering
\footnotesize
\setlength{\tabcolsep}{3pt}
\begin{tabular}{lrrr}
\toprule
Density bin & FineWeb-2 & FinePDFs & FineWiki \\
\midrule
$[0.00, 0.05)$ & 46.9\% & 57.3\% & 41.3\% \\
$[0.05, 0.10)$ & 21.9\% & 17.4\% & 11.3\% \\
$[0.10, 0.20)$ & 21.9\% & 16.5\% & 15.7\% \\
$[0.20, 0.30)$ & 7.0\%  & 6.1\%  & 15.7\% \\
$[0.30, 1.00)$ & 2.3\%  & 2.7\%  & 16.0\% \\
\bottomrule
\end{tabular}
\caption{Medical-term-density distribution (\%) per corpus on the prefiltered medical subset.}
\label{tab:density-distribution}
\end{table}

\subsection{Joint Distribution Across Axes}
\label{app:joint-dist}

\begin{figure*}[t]
\centering
\includegraphics[width=\linewidth]{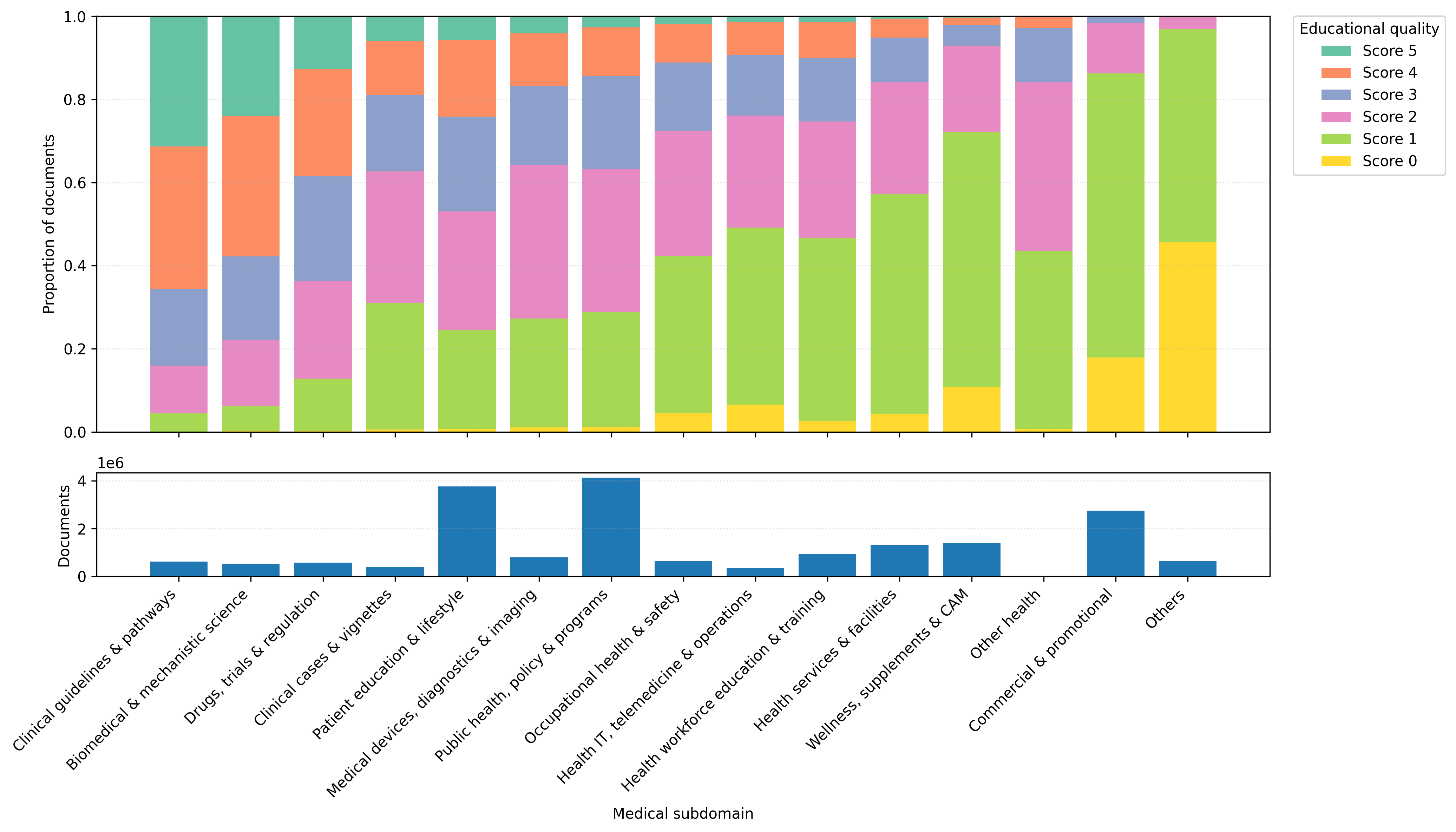}
\caption{Educational-quality score distribution per subdomain on the FineWeb-2 medical subset (stacked bars, normalized by subdomain document count). Document counts per subdomain shown below.}
\label{fig:joint-edu-subdomain}
\end{figure*}

\begin{figure}[t]
\centering
\includegraphics[width=\linewidth]{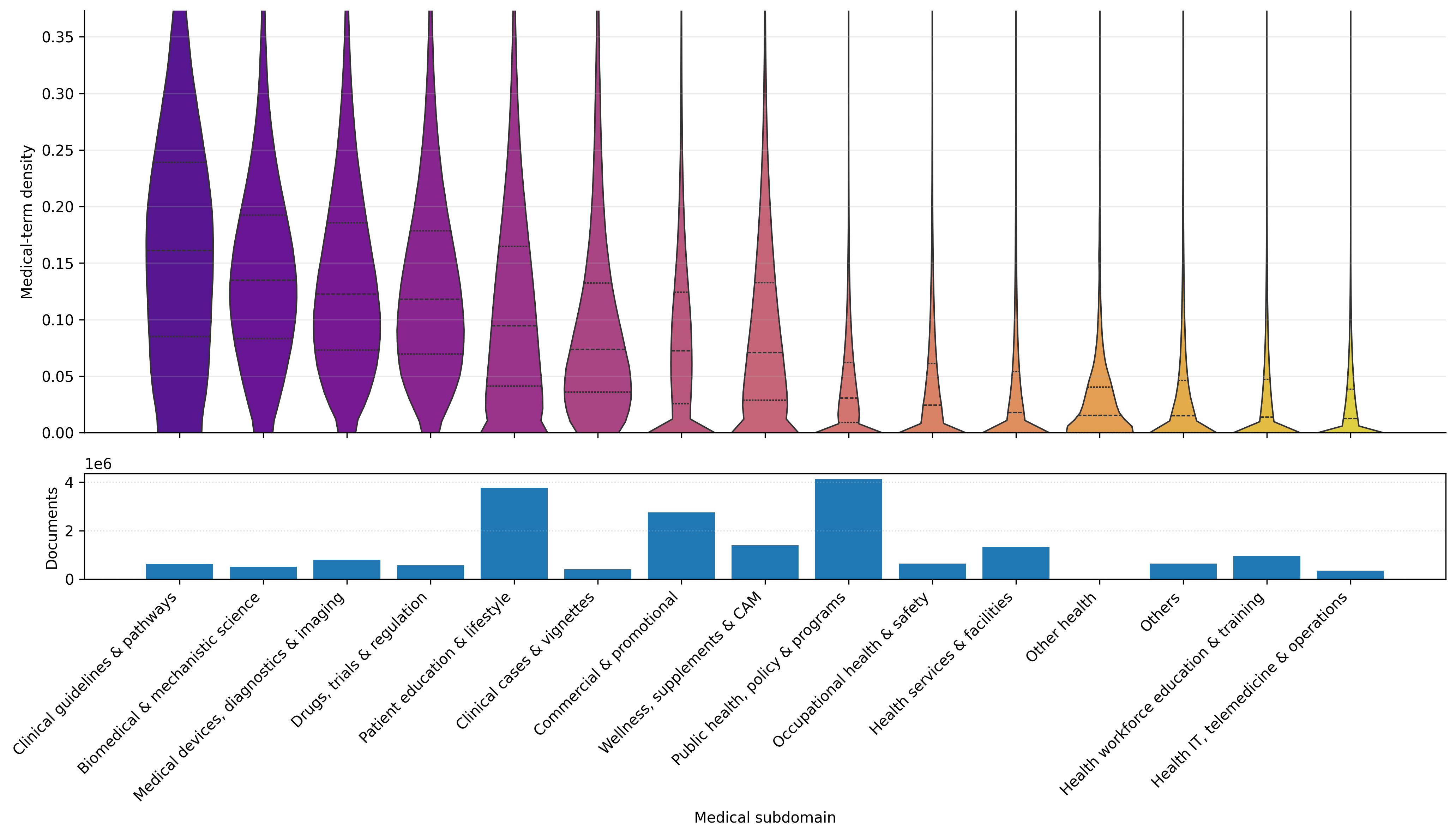}
\caption{Medical-term-density distributions per subdomain on the FineWeb-2 medical subset (violin). Document counts per subdomain shown below.}
\label{fig:joint-density-subdomain}
\end{figure}

\begin{figure}[t]
\centering
\includegraphics[width=\linewidth]{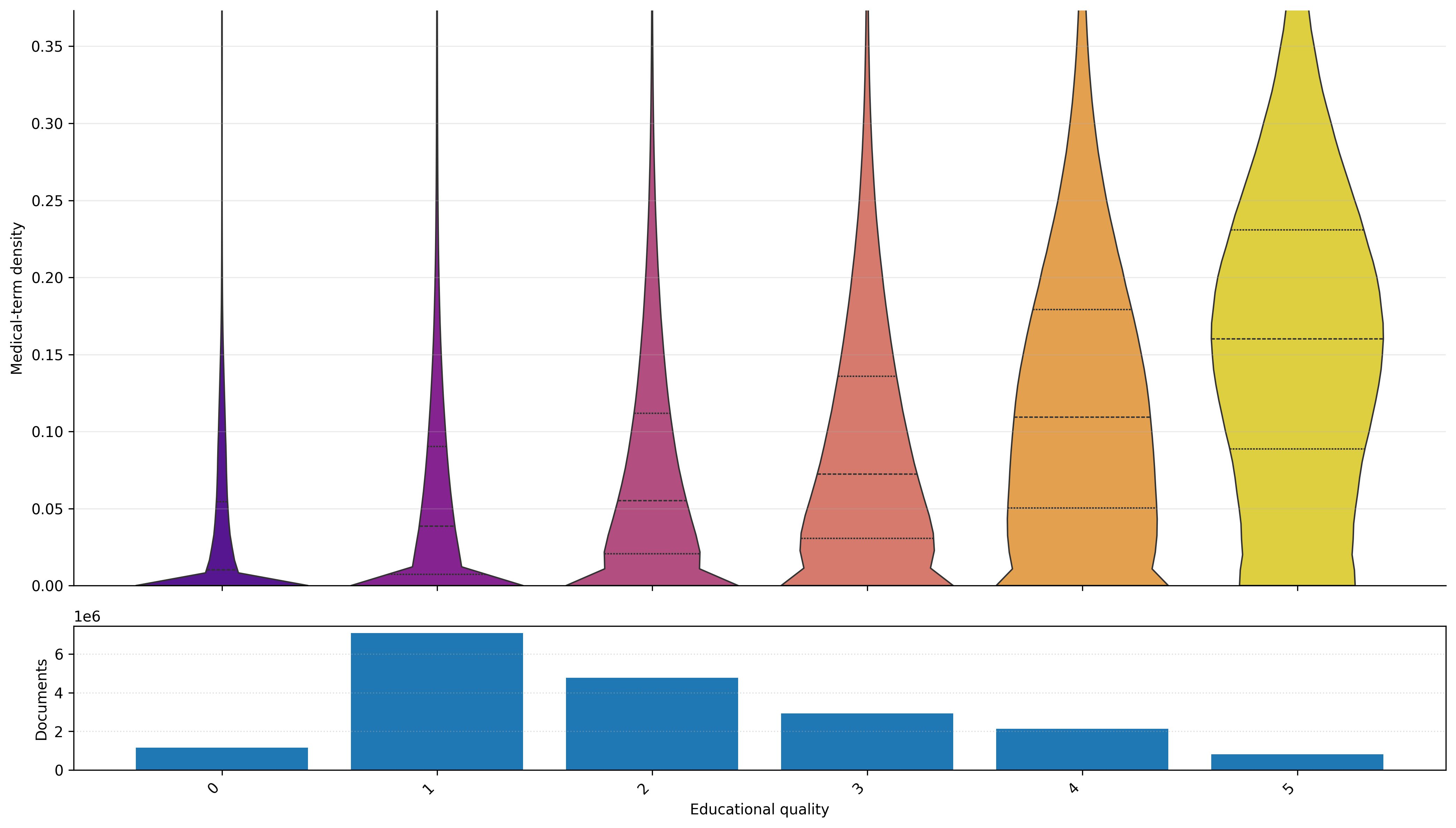}
\caption{Medical-term-density distributions per edu-quality score on the FineWeb-2 medical subset (violin). Density rises with edu, but within-bin spread is wide at high edu.}
\label{fig:joint-density-edu}
\end{figure}

\paragraph{Joint distribution.}
All three figures use the FineWeb-2 medical subset.
Figure~\ref{fig:joint-edu-subdomain} shows the educational-quality distribution per subdomain (stacked bar).
Figures~\ref{fig:joint-density-subdomain} and~\ref{fig:joint-density-edu} show medical-term-density distributions per subdomain and per edu score, respectively (violin); at high edu (4--5), the within-edu density spread is wide --- density and edu carry complementary information.

\subsection{Per-Task Filtering Ablation}
\label{app:per-task}

\paragraph{Training setup.}
To keep ablation training simple, we run single-stage pretraining at 1{,}024-token context; documents with fewer than 10 words or in the \emph{Others} subdomain are dropped.

\paragraph{Per-task scores.}
Table~\ref{tab:filter-per-task} reports per-task scores for all rows in Table~\ref{tab:filter-ablation}. No single configuration wins every task.

\paragraph{Bio\&Cli composition.}
\label{app:bio-cli}
The \emph{Bio\&Cli} filter targets the five biomedical/clinical subdomains we expect to be most relevant for medical-encoder pretraining: \emph{Biomedical \& mechanistic science}, \emph{Clinical cases \& vignettes}, \emph{Clinical guidelines \& pathways}, \emph{Drugs, trials \& regulation}, and \emph{Medical devices, diagnostics \& imaging}.

\begin{table*}[t]
\centering
\scriptsize
\setlength{\tabcolsep}{4pt}
\begin{tabular}{l r r@{}l r@{}l r@{}l r@{}l r@{}l r@{}l r@{}l}
\toprule
 & & \multicolumn{4}{c}{QUAERO} & \multicolumn{4}{c}{E3C} & \multicolumn{2}{c}{MORFITT} & \multicolumn{2}{c}{DEFT2021} & \multicolumn{2}{c}{DIAMED} \\
\cmidrule(lr){3-6} \cmidrule(lr){7-10} \cmidrule(lr){11-12} \cmidrule(lr){13-14} \cmidrule(lr){15-16}
Configuration & \#Words & \multicolumn{2}{c}{EMEA} & \multicolumn{2}{c}{MEDLINE} & \multicolumn{2}{c}{CLINICAL} & \multicolumn{2}{c}{TEMPORAL} & \multicolumn{2}{c}{CLS} & \multicolumn{2}{c}{NER} & \multicolumn{2}{c}{CLS} \\
\midrule
\multicolumn{16}{l}{\emph{Source baselines}} \\
NACHOS & 1.3B & 62.10 & ${}_{\pm 1.84}$ & 57.82 & ${}_{\pm 0.34}$ & 58.96 & ${}_{\pm 0.33}$ & 82.82 & ${}_{\pm 1.13}$ & 69.84 & ${}_{\pm 0.65}$ & 61.48 & ${}_{\pm 0.35}$ & 65.60 & ${}_{\pm 1.57}$ \\
TransCorpus-bio-fr & 5.2B & 61.52 & ${}_{\pm 1.10}$ & 55.65 & ${}_{\pm 0.78}$ & 59.57 & ${}_{\pm 1.01}$ & 83.30 & ${}_{\pm 0.55}$ & 70.58 & ${}_{\pm 0.81}$ & 60.33 & ${}_{\pm 1.16}$ & 65.76 & ${}_{\pm 1.58}$ \\
FW2-Med & 7.2B & \textbf{66.53} & ${}_{\pm 0.61}$ & 57.13 & ${}_{\pm 0.69}$ & 56.60 & ${}_{\pm 1.01}$ & 83.00 & ${}_{\pm 1.12}$ & \underline{71.12} & ${}_{\pm 0.20}$ & 61.18 & ${}_{\pm 0.81}$ & 56.70 & ${}_{\pm 2.88}$ \\
\midrule
\multicolumn{16}{l}{\emph{FW2-Med: single-axis filters}} \\
Bio\&Cli & 1.2B & 65.06 & ${}_{\pm 1.64}$ & 57.81 & ${}_{\pm 0.59}$ & 59.49 & ${}_{\pm 0.90}$ & 82.87 & ${}_{\pm 0.77}$ & \textbf{71.29} & ${}_{\pm 0.37}$ & 61.01 & ${}_{\pm 0.57}$ & 63.53 & ${}_{\pm 1.38}$ \\
Edu $\geq 2$ & 4.7B & 65.40 & ${}_{\pm 1.49}$ & 56.56 & ${}_{\pm 0.58}$ & 57.92 & ${}_{\pm 1.15}$ & 83.62 & ${}_{\pm 0.78}$ & 69.59 & ${}_{\pm 0.78}$ & 61.02 & ${}_{\pm 0.68}$ & 64.61 & ${}_{\pm 5.23}$ \\
Edu $\geq 4$ & 1.6B & 65.29 & ${}_{\pm 0.58}$ & 57.06 & ${}_{\pm 0.68}$ & 58.63 & ${}_{\pm 1.41}$ & 83.78 & ${}_{\pm 0.50}$ & 70.33 & ${}_{\pm 0.62}$ & 60.53 & ${}_{\pm 0.68}$ & 65.58 & ${}_{\pm 3.44}$ \\
Med-term $\geq 0.1$ & 2.5B & 65.38 & ${}_{\pm 1.25}$ & 57.64 & ${}_{\pm 0.69}$ & 58.86 & ${}_{\pm 1.08}$ & \textbf{83.82} & ${}_{\pm 0.43}$ & 70.39 & ${}_{\pm 0.91}$ & 61.84 & ${}_{\pm 0.97}$ & 67.43 & ${}_{\pm 4.53}$ \\
Med-term $\geq 0.2$ & 762M & 64.77 & ${}_{\pm 0.57}$ & 56.55 & ${}_{\pm 1.23}$ & \textbf{59.65} & ${}_{\pm 0.85}$ & 83.64 & ${}_{\pm 0.56}$ & 69.95 & ${}_{\pm 0.66}$ & \textbf{62.20} & ${}_{\pm 0.52}$ & \underline{69.77} & ${}_{\pm 2.65}$ \\
\midrule
\multicolumn{16}{l}{\emph{FW2-Med: intersection combinations}} \\
Bio\&Cli $\cap$ Edu $\geq 4$ & 587M & 65.68 & ${}_{\pm 0.39}$ & 56.87 & ${}_{\pm 0.63}$ & 59.57 & ${}_{\pm 1.29}$ & 83.48 & ${}_{\pm 0.50}$ & 70.45 & ${}_{\pm 0.70}$ & 60.84 & ${}_{\pm 0.52}$ & 65.54 & ${}_{\pm 1.71}$ \\
Bio\&Cli $\cap$ Med-term $\geq 0.1$ & 702M & \underline{66.20} & ${}_{\pm 0.67}$ & 57.66 & ${}_{\pm 0.35}$ & 58.40 & ${}_{\pm 0.88}$ & \underline{83.80} & ${}_{\pm 0.31}$ & 70.90 & ${}_{\pm 1.09}$ & 61.43 & ${}_{\pm 0.86}$ & 66.79 & ${}_{\pm 3.19}$ \\
Bio\&Cli $\cap$ Med-term $\geq 0.2$ & 264M & 64.94 & ${}_{\pm 1.29}$ & 57.42 & ${}_{\pm 0.30}$ & \underline{59.59} & ${}_{\pm 1.99}$ & 82.82 & ${}_{\pm 0.48}$ & 71.01 & ${}_{\pm 0.40}$ & 60.43 & ${}_{\pm 0.91}$ & \textbf{70.63} & ${}_{\pm 1.65}$ \\
Edu $\geq 4$ $\cap$ Med-term $\geq 0.1$ & 933M & 64.77 & ${}_{\pm 1.37}$ & \textbf{58.07} & ${}_{\pm 1.01}$ & 59.06 & ${}_{\pm 1.32}$ & 83.64 & ${}_{\pm 0.44}$ & 70.94 & ${}_{\pm 0.70}$ & 61.75 & ${}_{\pm 0.75}$ & 68.67 & ${}_{\pm 2.31}$ \\
\midrule
\multicolumn{16}{l}{\emph{FW2-Med: union combinations}} \\
Bio\&Cli $\cup$ Edu $\geq 4$ & 2.2B & 64.15 & ${}_{\pm 1.00}$ & 57.30 & ${}_{\pm 0.49}$ & 58.39 & ${}_{\pm 0.82}$ & 83.76 & ${}_{\pm 0.57}$ & 69.95 & ${}_{\pm 0.74}$ & 61.17 & ${}_{\pm 0.83}$ & 66.90 & ${}_{\pm 1.58}$ \\
Bio\&Cli $\cup$ Med-term $\geq 0.1$ & 3.0B & 65.23 & ${}_{\pm 1.07}$ & \underline{57.99} & ${}_{\pm 0.31}$ & 58.91 & ${}_{\pm 1.16}$ & 81.89 & ${}_{\pm 0.81}$ & 69.88 & ${}_{\pm 0.53}$ & 59.84 & ${}_{\pm 0.76}$ & 65.80 & ${}_{\pm 4.94}$ \\
Edu $\geq 4$ $\cup$ Med-term $\geq 0.1$ & 3.2B & 64.15 & ${}_{\pm 1.90}$ & 57.55 & ${}_{\pm 0.54}$ & 59.51 & ${}_{\pm 1.48}$ & 83.62 & ${}_{\pm 0.57}$ & 69.19 & ${}_{\pm 1.00}$ & \underline{61.93} & ${}_{\pm 0.27}$ & 65.06 & ${}_{\pm 1.65}$ \\
\bottomrule
\end{tabular}
\caption{Per-task scores and corpus sizes for the filter-axis ablation: mean$\pm$std F1 over multiple seeds on the test split. Best per task in bold, second best underlined. Tasks: QUAERO EMEA/MEDLINE NER, E3C clinical/temporal NER, MORFITT classification, DEFT2021 NER, DIAMED classification.}
\label{tab:filter-per-task}
\end{table*}

\section{Signal-Amplifying Rephrasing}
\label{app:rephrasing}

\subsection{Rephrasing Prompts}
\label{app:rephrasing-prompts}

Prompts~\ref{app:rephrase-stage1-prompt} and~\ref{app:rephrase-stage2-prompt} show the stage-1 and stage-2 prompts. Stage 1 proposes (genre, audience) pairs via \emph{brainstorm-then-couple}. Stage 2 rephrases the source under one such pair, with faithful densification, PII handling, and surface variation (register and abbreviation density).

\subsection{Rephrased-Output Post-Processing}
\label{app:rephrasing-postproc}

LLM rephrasing can fail in ways that produce unusable outputs: language drift, degenerate repetition, or malformed text. We filter these out using DataTrove~\citep{penedo2024datatrove}:
\begin{itemize}\itemsep0pt
  \item \textbf{Language filter}: drop documents whose target-language confidence is below 0.5.
  \item \textbf{Gopher repetition filter}~\citep{rae_scaling_2022}: drop documents with anomalous repetition, n-gram patterns, or symbol-to-word ratios.
\end{itemize}

\subsection{Preliminary Rephraser Benchmark}
\label{app:rephrasing-llmaj}

Training-time rephraser evaluation requires one pretraining run per candidate. To shortlist before this step, we screen a broader candidate pool on a 10k FineWeb-2 sub-sample with low-compute proxies:
\begin{itemize}\itemsep0pt
  \item \emph{Inference time}: wall-clock to rephrase 10k documents on $4\times$H100. Bounds the operational cost of corpus-scale rephrasing.
  \item \emph{Compression ratio}: rephrased-words / source-words. Flags over-compression (collapsing to a summary) versus acceptable densification.
  \item \emph{Medical-term density} on the rephrasings. Signals whether the rephraser preserves the source's entity-rich content rather than diluting it.
  \item \emph{Factuality}, \emph{faithfulness}, and \emph{style adherence}: each scored 0--5 by an LLM-as-judge (Intellect-3~\citep{primeintellect_intellect3_2025} with chain-of-thought reasoning, held out from the candidate pool). The three flag invented facts, loss of medical content, and genre/style violations respectively.
\end{itemize}
Compression, density, and the three judge scores are computed on each rephraser's top 5000 rephrasings by medical-term density, so each model is evaluated on its high-density tail. Table~\ref{tab:rephraser-bench} reports the benchmark. Asterisks (*) mark the shortlist taken to training-time evaluation.

\begin{table*}[t]
\centering
\scriptsize
\setlength{\tabcolsep}{5pt}
\begin{tabular}{lrccccc}
\toprule
Rephraser & Time & Cmp.\ ratio & Med density & Factuality & Faithfulness & Style \\
\midrule
Qwen3-30B-A3B-Instruct               &  9\,min & 0.72 & 0.19 & 4.44 & 4.36 & 4.24 \\
Qwen3-30B-A3B-Instruct (thinking)    & 51\,min & 0.32 & 0.22 & 4.69 & 3.84 & 4.19 \\
Qwen3-Next-80B-A3B               & 16\,min & 0.65 & 0.20 & 4.40 & 4.30 & 4.11 \\
Qwen3-235B-A22B-Instruct         & 65\,min & 0.81 & 0.20 & 4.46 & 4.43 & 4.36 \\
Qwen3.5-9B                           & 10\,min & 0.54 & 0.19 & 4.65 & 4.32 & 4.34 \\
Qwen3.5-27B                          & 28\,min & 0.54 & 0.22 & 4.77 & 4.54 & 4.54 \\
Qwen3.5-35B-A3B~\citep{qwen35_2026}*                 & 14\,min & 0.58 & 0.21 & 4.71 & 4.52 & 4.46 \\
Qwen3.5-122B-A10B*               & 25\,min & 0.66 & 0.20 & 4.77 & 4.54 & 4.33 \\
MedGemma-27B~\citep{sellergren_medgemma_2025}*                        & 22\,min & 0.55 & 0.18 & 4.67 & 4.28 & 4.17 \\
Gemma-4-26B-A4B~\citep{gemma4_2025}*                     & 11\,min & 0.55 & 0.25 & 4.72 & 4.19 & 4.32 \\
Gemma-4-31B                          & 31\,min & 0.45 & 0.26 & 4.73 & 4.18 & 4.48 \\
GPT-OSS-20B (low reasoning)          & 11\,min & 0.55 & 0.23 & 3.50 & 3.55 & 3.89 \\
GPT-OSS-20B (med.\ reasoning)        & 73\,min & 0.59 & 0.22 & 3.62 & 3.68 & 4.04 \\
GPT-OSS-120B (low reasoning)~\citep{openai2025gptoss120bgptoss20bmodel}*    & 13\,min & 0.57 & 0.25 & 4.25 & 4.22 & 4.45 \\
GPT-OSS-120B (med.\ reasoning)       & 45\,min & 0.60 & 0.25 & 4.27 & 4.24 & 4.49 \\
Ministral-3-14B-Instruct~\citep{liu_ministral_2026}             & 37\,min & 1.02 & 0.09 & 1.88 & 1.94 & 2.05 \\
Mistral-Small-3.2-24B~\citep{mistral_small_2025}                & 13\,min & 0.40 & 0.19 & 3.66 & 3.23 & 3.56 \\
Nemotron-3-Nano-30B-A3B~\citep{nvidia_nemotron_2025}          & 10\,min & 0.59 & 0.14 & 2.75 & 2.49 & 2.81 \\
GLM-4.7-Flash~\citep{zai_glm47_2025}                        & 54\,min & 0.42 & 0.21 & 4.18 & 3.62 & 3.92 \\
\bottomrule
\end{tabular}
\caption{Rephraser benchmark on a 10k FineWeb-2 sub-sample. Asterisks (*) mark rephrasers taken to training-time evaluation (Table~\ref{tab:rephrase-100k}).}
\label{tab:rephraser-bench}
\end{table*}

\subsection{Per-Task Rephrasing Ablation}
\label{app:per-task-rephrase}

Tables~\ref{tab:rephrase-100k-per-task} and~\ref{tab:rephrase-1M-per-task} report per-task scores for the 100k- and 1M-source-document rephrasing ablations (aggregate rows in Tables~\ref{tab:rephrase-100k} and~\ref{tab:rephrase-20b}, respectively).

\begin{table*}[t]
\centering
\scriptsize
\setlength{\tabcolsep}{4pt}
\begin{tabular}{l r r@{}l r@{}l r@{}l r@{}l r@{}l r@{}l r@{}l}
\toprule
 & & \multicolumn{4}{c}{QUAERO} & \multicolumn{4}{c}{E3C} & \multicolumn{2}{c}{MORFITT} & \multicolumn{2}{c}{DEFT2021} & \multicolumn{2}{c}{DIAMED} \\
\cmidrule(lr){3-6} \cmidrule(lr){7-10} \cmidrule(lr){11-12} \cmidrule(lr){13-14} \cmidrule(lr){15-16}
Configuration & \#Words & \multicolumn{2}{c}{EMEA} & \multicolumn{2}{c}{MEDLINE} & \multicolumn{2}{c}{CLINICAL} & \multicolumn{2}{c}{TEMPORAL} & \multicolumn{2}{c}{CLS} & \multicolumn{2}{c}{NER} & \multicolumn{2}{c}{CLS} \\
\midrule
\multicolumn{16}{l}{\emph{Baseline}} \\
Raw (no rephrasing) & 38M & 45.17 & ${}_{\pm 0.89}$ & 33.61 & ${}_{\pm 0.67}$ & 50.18 & ${}_{\pm 0.70}$ & 74.32 & ${}_{\pm 0.46}$ & 58.60 & ${}_{\pm 0.63}$ & 42.60 & ${}_{\pm 0.73}$ & \underline{53.17} & ${}_{\pm 1.63}$ \\
\midrule
\multicolumn{16}{l}{\emph{Standard MGA}} \\
Qwen3.5-35B-A3B & 50M & 41.64 & ${}_{\pm 1.36}$ & 30.17 & ${}_{\pm 1.55}$ & 47.93 & ${}_{\pm 1.75}$ & 74.01 & ${}_{\pm 1.09}$ & 53.83 & ${}_{\pm 0.83}$ & 38.00 & ${}_{\pm 1.54}$ & 52.58 & ${}_{\pm 3.58}$ \\
\midrule
\multicolumn{16}{l}{\emph{Our recipe, varying the rephraser}} \\
Qwen3.5-35B-A3B & 15M & \textbf{55.13} & ${}_{\pm 2.17}$ & \textbf{45.17} & ${}_{\pm 0.86}$ & 53.29 & ${}_{\pm 1.73}$ & \textbf{79.32} & ${}_{\pm 0.97}$ & \textbf{62.89} & ${}_{\pm 1.16}$ & \textbf{50.31} & ${}_{\pm 0.41}$ & \textbf{54.90} & ${}_{\pm 2.30}$ \\
Qwen3.5-122B-A10B & 16M & \underline{52.64} & ${}_{\pm 0.53}$ & \underline{44.30} & ${}_{\pm 0.56}$ & \underline{53.73} & ${}_{\pm 1.58}$ & \underline{78.96} & ${}_{\pm 0.37}$ & \underline{62.49} & ${}_{\pm 0.63}$ & 48.59 & ${}_{\pm 0.81}$ & 50.05 & ${}_{\pm 2.69}$ \\
Gemma-4-26B-A4B & 16M & 48.34 & ${}_{\pm 1.18}$ & 43.05 & ${}_{\pm 0.66}$ & \textbf{54.39} & ${}_{\pm 0.63}$ & 76.49 & ${}_{\pm 0.71}$ & 61.64 & ${}_{\pm 0.36}$ & \underline{49.02} & ${}_{\pm 0.26}$ & 50.51 & ${}_{\pm 2.14}$ \\
MedGemma-27B & 21M & 42.32 & ${}_{\pm 0.99}$ & 28.81 & ${}_{\pm 0.60}$ & 48.26 & ${}_{\pm 1.78}$ & 74.03 & ${}_{\pm 1.67}$ & 53.33 & ${}_{\pm 1.45}$ & 41.39 & ${}_{\pm 0.87}$ & 42.15 & ${}_{\pm 2.70}$ \\
GPT-OSS-120B & 17M & 38.10 & ${}_{\pm 1.96}$ & 26.98 & ${}_{\pm 1.23}$ & 49.00 & ${}_{\pm 0.61}$ & 70.52 & ${}_{\pm 0.67}$ & 53.18 & ${}_{\pm 1.53}$ & 34.91 & ${}_{\pm 1.85}$ & 36.14 & ${}_{\pm 6.85}$ \\
\bottomrule
\end{tabular}
\caption{Per-task scores and corpus sizes for the 100k-source-document rephrasing ablation (rows in Table~\ref{tab:rephrase-100k}): mean$\pm$std F1 over multiple seeds on the test split. Best per task in bold, second best underlined.}
\label{tab:rephrase-100k-per-task}
\end{table*}

\begin{table*}[t]
\centering
\scriptsize
\setlength{\tabcolsep}{4pt}
\begin{tabular}{l r r@{}l r@{}l r@{}l r@{}l r@{}l r@{}l r@{}l}
\toprule
 & & \multicolumn{4}{c}{QUAERO} & \multicolumn{4}{c}{E3C} & \multicolumn{2}{c}{MORFITT} & \multicolumn{2}{c}{DEFT2021} & \multicolumn{2}{c}{DIAMED} \\
\cmidrule(lr){3-6} \cmidrule(lr){7-10} \cmidrule(lr){11-12} \cmidrule(lr){13-14} \cmidrule(lr){15-16}
Configuration & \#Words & \multicolumn{2}{c}{EMEA} & \multicolumn{2}{c}{MEDLINE} & \multicolumn{2}{c}{CLINICAL} & \multicolumn{2}{c}{TEMPORAL} & \multicolumn{2}{c}{CLS} & \multicolumn{2}{c}{NER} & \multicolumn{2}{c}{CLS} \\
\midrule
\multicolumn{16}{l}{\emph{Baseline}} \\
Raw (no rephrasing) & 392M & \textbf{65.49} & ${}_{\pm 1.73}$ & 56.19 & ${}_{\pm 0.47}$ & \textbf{59.78} & ${}_{\pm 2.15}$ & 82.99 & ${}_{\pm 0.55}$ & 68.98 & ${}_{\pm 1.04}$ & 59.94 & ${}_{\pm 0.77}$ & 64.89 & ${}_{\pm 2.26}$ \\
\midrule
\multicolumn{16}{l}{\emph{Rephrased only}} \\
Qwen & 158M & 63.14 & ${}_{\pm 1.25}$ & 54.76 & ${}_{\pm 1.14}$ & 58.58 & ${}_{\pm 1.00}$ & 82.97 & ${}_{\pm 0.50}$ & \underline{71.61} & ${}_{\pm 0.64}$ & 61.21 & ${}_{\pm 0.45}$ & 64.47 & ${}_{\pm 1.64}$ \\
Gemma & 158M & 63.20 & ${}_{\pm 0.47}$ & 54.69 & ${}_{\pm 0.40}$ & 57.55 & ${}_{\pm 2.76}$ & 82.94 & ${}_{\pm 0.61}$ & 70.84 & ${}_{\pm 1.35}$ & 60.02 & ${}_{\pm 0.39}$ & 66.28 & ${}_{\pm 1.73}$ \\
Qwen + Gemma (1:1) & 158M & 61.40 & ${}_{\pm 0.91}$ & 55.18 & ${}_{\pm 1.03}$ & 56.77 & ${}_{\pm 1.02}$ & 82.05 & ${}_{\pm 0.59}$ & 70.36 & ${}_{\pm 0.43}$ & \textbf{62.18} & ${}_{\pm 0.82}$ & \textbf{67.97} & ${}_{\pm 2.49}$ \\
Qwen, density-filtered & 90M & 62.85 & ${}_{\pm 1.24}$ & 54.55 & ${}_{\pm 1.21}$ & 56.82 & ${}_{\pm 0.54}$ & 81.87 & ${}_{\pm 0.59}$ & 70.59 & ${}_{\pm 0.46}$ & 59.95 & ${}_{\pm 0.51}$ & 67.46 & ${}_{\pm 1.43}$ \\
\midrule
\multicolumn{16}{l}{\emph{Rephrased + raw}} \\
Qwen + raw & 550M & \underline{64.95} & ${}_{\pm 1.10}$ & \underline{57.01} & ${}_{\pm 0.54}$ & 57.94 & ${}_{\pm 0.74}$ & \underline{83.21} & ${}_{\pm 0.52}$ & 69.57 & ${}_{\pm 1.22}$ & \underline{62.00} & ${}_{\pm 0.63}$ & 65.92 & ${}_{\pm 3.48}$ \\
Qwen + filtered raw & 211M & 63.09 & ${}_{\pm 1.72}$ & \textbf{57.56} & ${}_{\pm 0.42}$ & 58.10 & ${}_{\pm 0.81}$ & \textbf{83.88} & ${}_{\pm 0.57}$ & \textbf{71.71} & ${}_{\pm 0.76}$ & 61.20 & ${}_{\pm 0.66}$ & 66.95 & ${}_{\pm 2.68}$ \\
\quad Qwen:filtered raw = 2:1 & 157M & 59.73 & ${}_{\pm 1.62}$ & 55.50 & ${}_{\pm 0.81}$ & 55.06 & ${}_{\pm 0.75}$ & 82.62 & ${}_{\pm 0.61}$ & 71.35 & ${}_{\pm 0.75}$ & 60.68 & ${}_{\pm 0.25}$ & \underline{67.73} & ${}_{\pm 2.49}$ \\
\quad Qwen:filtered raw = 1:1 & 105M & 64.49 & ${}_{\pm 1.04}$ & 55.09 & ${}_{\pm 0.65}$ & \underline{59.02} & ${}_{\pm 0.71}$ & 82.52 & ${}_{\pm 0.48}$ & 69.89 & ${}_{\pm 0.64}$ & 61.15 & ${}_{\pm 0.81}$ & 66.11 & ${}_{\pm 0.94}$ \\
\quad Qwen:filtered raw = 1:2 & 79M & 61.87 & ${}_{\pm 2.47}$ & 51.83 & ${}_{\pm 0.64}$ & 56.15 & ${}_{\pm 0.75}$ & 82.99 & ${}_{\pm 0.37}$ & 69.63 & ${}_{\pm 0.68}$ & 59.45 & ${}_{\pm 0.39}$ & 61.73 & ${}_{\pm 3.28}$ \\
\bottomrule
\end{tabular}
\caption{Per-task scores and corpus sizes for the 1M-source-document rephrasing ablation (rows in Table~\ref{tab:rephrase-20b}): mean$\pm$std F1 over multiple seeds on the test split. Qwen and Gemma denote Qwen3.5-35B-A3B and Gemma-4-26B-A4B. Best per task in bold, second best underlined.}
\label{tab:rephrase-1M-per-task}
\end{table*}

\subsection{Medical-Content Gate Proxy}
\label{app:rephrasing-proxy}

The stage-1 gate rejects roughly 27.5\% of documents as non-medical in the 1M-scale ablation, yet each rejected document incurs one LLM call.
To avoid this cost, we approximate the gate with our precomputed upstream annotations (subdomain, edu, density) and evaluate candidate filters against the rephraser's judgment (Table~\ref{tab:has-medical-proxy}).
We adopt \emph{edu} $\geq 1 \land$ \emph{density} $\geq 0.01$ as a coarse pre-screen.
It recovers 89\% of the LLM gate's medical documents and saves 23.1\% of stage-1 LLM calls.

\begin{table}[t]
\centering
\footnotesize
\setlength{\tabcolsep}{4pt}
\begin{tabular}{lrrrr}
\toprule
Filter & Kept\% & P & R & F1 \\
\midrule
\multicolumn{5}{l}{\emph{Edu threshold}} \\
\emph{edu} $\geq 1$                                      & 93.9\% & 0.77 & 0.99 & \textbf{0.865} \\
\emph{edu} $\geq 2$                                      & 56.7\% & 0.93 & 0.73 & 0.818 \\
\midrule
\multicolumn{5}{l}{\emph{Density threshold}} \\
\emph{density} $\geq 0.01$                               & 80.0\% & 0.81 & 0.90 & 0.852 \\
\emph{density} $\geq 0.02$                               & 72.4\% & 0.83 & 0.83 & 0.831 \\
\emph{density} $\geq 0.05$                               & 53.3\% & 0.87 & 0.64 & 0.737 \\
\emph{density} $\geq 0.10$                               & 31.6\% & 0.91 & 0.40 & 0.550 \\
\midrule
\multicolumn{5}{l}{\emph{Subdomain whitelist}} \\
\emph{domain} $\in$ Bio\&Cli (\S\ref{app:subdomain-classifier})        & 15.7\% & 0.92 & 0.20 & 0.329 \\
\midrule
\multicolumn{5}{l}{\emph{Combined}} \\
\emph{density} $\geq 0.01 \land$ \emph{edu} $\geq 1$     & 76.9\% & 0.84 & 0.89 & \underline{0.862} \\
\bottomrule
\end{tabular}
\caption{Proxy filters built from precomputed multi-axis annotations to approximate the rephraser's medical-content gate. P/R/F1 against the rephraser judgment on 1M rows.}
\label{tab:has-medical-proxy}
\end{table}

\subsection{Faithfulness Audit}
\label{app:faithfulness-audit}

We audit rephrasing faithfulness on 501 (source, rephrased) pairs from the released corpus, sampled uniformly per source (167 each from FineWeb-2, FinePDFs, FineWiki).
An LLM judge (Qwen3.8-27B) assigns three binary flags per pair: \emph{invented facts} (a specific claim about the case unsupported by the source), \emph{altered negation}, and \emph{changed clinical relations}.
Table~\ref{tab:faithfulness-audit} reports the flag rates.

\begin{table}[t]
\centering
\footnotesize
\begin{tabular}{l r}
\toprule
Category & Rate (\%) \\
\midrule
Invented facts              & 9.0  \\
Altered negation            & 1.6  \\
Changed clinical relations  & 3.6  \\
Any of the three            & 10.4 \\
\bottomrule
\end{tabular}
\caption{Faithfulness audit: judge flag rates over 501 released (source, rephrased) pairs.}
\label{tab:faithfulness-audit}
\end{table}

We manually reviewed every pair flagged for invented facts: most flags are genuine inventions, and the rest are judge errors (benign glosses of concepts already in the source, allowed anonymization of blank template fields, ambiguous cases).
Two failure modes recur.
On thin or noisy sources that pass the medical-content gate, the rewriter can fabricate a fluent clinical narrative.
Hedged source statements can also be upgraded to confident claims (certainty inflation), including one polarity reversal on a prescribing guideline.

\section{Evaluation}
\label{app:evaluation}

This appendix details the evaluation protocols for DrBenchmark (\S\ref{sec:exp}) and the proprietary clinical NER task (\S\ref{sec:final-eval}).

\subsection{DrBenchmark Adaptation}
\label{app:adapted-benchmark}

The three adaptations each address a distinct failure mode of vanilla evaluation: HPO + multi-seed reruns control seed noise on small splits, task filtering drops noisy tasks, and cross-task aggregation produces a comparable per-model number.

\paragraph{HPO + multi-seed reruns.}
\citet{knafou_transbert_2025}'s adapted DrBenchmark protocol introduced HPO, 5-fold cross-validation, and deduplication. We adopt HPO and deduplication but replace cross-validation with multi-seed reruns on the dataset's built-in train/val/test split. BERT finetuning is stochastic, so seed-averaging controls run-to-run noise; skipping CV also frees compute for more HPO trials. For each (model, task) pair we run HPO on the validation split with Ray Tune,\footnote{\url{https://docs.ray.io/en/latest/tune/index.html}} then re-train under 5 random seeds on the test split and report mean and standard deviation.

\paragraph{Task filtering.}
DrBenchmark aggregates tasks of uneven quality. We assess each task by data size, seed-to-seed noise, and correlation with the rest of the benchmark. To estimate the latter two, we run our HPO + multi-seed protocol on a small set of representative models and compute two metrics:
\begin{itemize}\itemsep1pt
  \item \emph{Signal-to-Noise Ratio} (SNR), the ratio of between-model variance to within-model seed variance:
    \[
      \mathrm{SNR} = \frac{\mathrm{Var}_m(\bar{s}_m)}{\mathrm{Mean}_m\!\left[\mathrm{Var}_{\text{seeds}}(s_m)\right]},
    \]
    where $\bar{s}_m$ is model $m$'s seed-averaged score. SNR captures whether model differences exceed seed noise.
  \item \emph{Average absolute Pearson correlation} ($\overline{|r|}$): for each task, the mean absolute Pearson correlation between its per-model mean scores and those of every other task. $\overline{|r|}$ captures how closely the task ranks models like the rest of the benchmark.
\end{itemize}

After filtering, seven tasks remain across five datasets: QUAERO (NER on EMEA and MEDLINE), E3C (clinical and temporal NER), MORFITT (specialty classification), DIAMED (diagnostic-category classification), and DEFT-2021 (NER).

\begin{figure}[t]
\centering
\includegraphics[width=\linewidth]{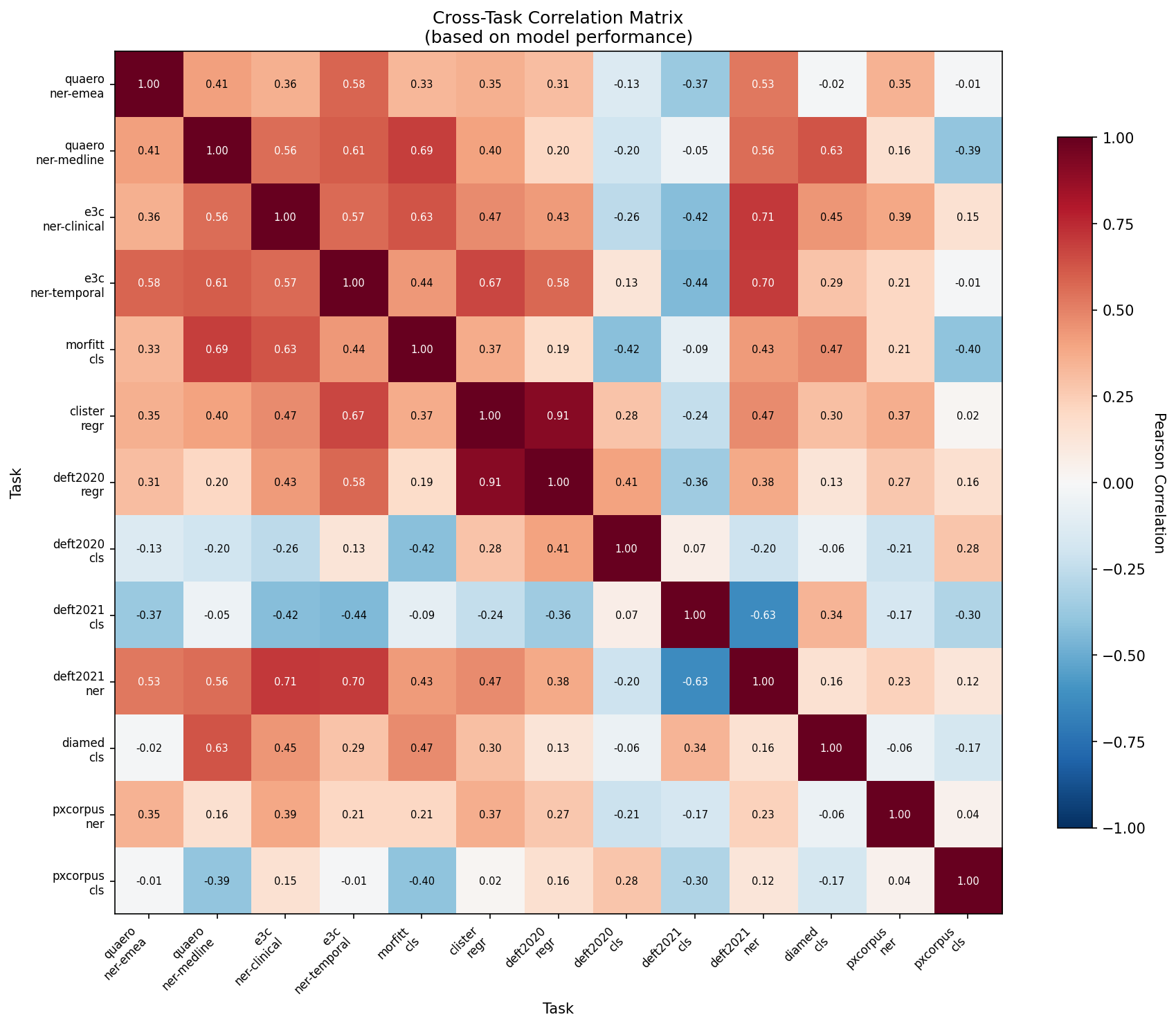}
\caption{Pairwise Pearson correlation between DrBenchmark tasks, computed over per-model mean scores. Cells with low absolute correlation against most others identify outlier tasks.}
\label{fig:drbenchmark-correlation}
\end{figure}

Table~\ref{tab:task-quality} reports per-task SNR, $\overline{|r|}$, and score range. Figure~\ref{fig:drbenchmark-correlation} shows the full cross-task correlation matrix.

\begin{table}[t]
\centering
\footnotesize
\setlength{\tabcolsep}{4pt}
\begin{tabular}{lccc}
\toprule
Task & SNR & $\overline{|r|}$ & Range \\
\midrule
\texttt{quaero-ner-emea}     & 1.74  & 0.31 & 0.062 \\
\texttt{quaero-ner-medline}  & 4.23  & 0.41 & 0.052 \\
\texttt{e3c-ner-clinical}    & 1.35  & 0.45 & 0.058 \\
\texttt{e3c-ner-temporal}    & 2.42  & 0.44 & 0.045 \\
\texttt{morfitt-cls}         & 1.89  & 0.39 & 0.059 \\
\texttt{clister-regr}        & 9.81  & 0.41 & 0.068 \\
\texttt{deft2020-regr}       & 9.12  & 0.36 & 0.045 \\
\texttt{deft2020-cls}        & 0.62  & 0.22 & 0.322 \\
\texttt{deft2021-cls}        & 11.04 & 0.29 & 0.301 \\
\texttt{deft2021-ner}        & 3.90  & 0.43 & 0.076 \\
\texttt{diamed-cls}          & 2.25  & 0.26 & 0.162 \\
\texttt{pxcorpus-ner}        & 0.82  & 0.22 & 0.010 \\
\texttt{pxcorpus-cls}        & 0.33  & 0.17 & 0.023 \\
\bottomrule
\end{tabular}
\caption{Per-task SNR, average absolute Pearson cross-task correlation, and score range across models.}
\label{tab:task-quality}
\end{table}

\paragraph{Cross-task aggregation.}
Tasks operate on different scales, so a plain mean over raw task scores lets high-magnitude tasks dominate. After averaging seeds for each (model, task) pair, we aggregate across tasks with two complementary per-model metrics, each reported as mean$\pm$SE over the 7 retained tasks:
\begin{itemize}\itemsep1pt
  \item \emph{Min-Max normalized scores}: for each task, normalize per-model scores into $[0,1]$ via $(x - \min)/(\max - \min)$ across the model set, then average across tasks and rescale to 0--100. Captures the magnitude of differences: a model that underperforms on a few tasks is penalized heavily.
  \item \emph{Win Probability}: for each ordered pair of models $(A, B)$, compute the fraction of tasks where $A$'s mean score exceeds $B$'s (ties counted as $0.5$); each model's reported value is the average of these fractions over all opponents, rescaled to 0--100. Captures rank consistency: robust to outliers but ignores effect size.
\end{itemize}

\subsection{Real-World Clinical NER Task}
\label{app:real-world-ner}

\paragraph{Taxonomy.}
The taxonomy comprises 12 entity classes describing medical concepts attached to the patient, and 9 qualifier classes scoping entities by modal markers (Table~\ref{tab:realworld-ner-taxonomy}).

\begin{table*}[h]
\centering
\footnotesize
\setlength{\tabcolsep}{4pt}
\begin{tabular}{ll}
\toprule
Label & Description \\
\midrule
\multicolumn{2}{l}{\emph{Entities (12 classes)}} \\
pathology              & Disease, syndrome, infection, or chronic condition       \\
drug\_based\_treatment & Pharmacological treatment by brand or generic name       \\
exam                   & Diagnostic or screening examination                      \\
biometry               & Patient biometric parameter                              \\
labs\_result           & Biological analyte or laboratory result                  \\
procedure              & Therapeutic or interventional procedure                  \\
lifestyle\_factor      & Lifestyle attribute (tobacco, alcohol, diet)             \\
vaccine                & Vaccine name                                             \\
allergy                & Allergic or intolerance condition                        \\
pregnancy\_status      & Pregnancy state or pregnancy-related status              \\
posology               & Drug dosage and administration schedule                  \\
score                  & Named clinical or biomedical score                       \\
\midrule
\multicolumn{2}{l}{\emph{Qualifiers (9 classes)}} \\
date\_duration                    & Date, time, or duration expression                \\
directive                         & Marker of intent or recommendation                \\
negation                          & Marker negating an entity                         \\
measurement\_result\_qualitative  & Qualitative descriptor of a measurement           \\
measurement\_result\_quantitative & Numerical measurement with unit                   \\
frequency                         & Frequency or periodicity expression               \\
uncertainty                       & Marker of doubt or uncertainty                    \\
family\_relationship              & Marker scoping to a family member                 \\
conditionality                    & Marker of conditional dependency                  \\
\bottomrule
\end{tabular}
\caption{Taxonomy of the proprietary clinical NER task: 12 entity classes and 9 qualifier classes.}
\label{tab:realworld-ner-taxonomy}
\end{table*}

\paragraph{Corpus statistics.}
Source documents are pseudonymized French clinical consultation summaries and short structured patient-record entries.
Each document is split into individual bullet points or single-line entries, with each forming one datapoint.
Table~\ref{tab:realworld-ner-stats} reports per-split statistics.

\begin{table}[h]
\centering
\footnotesize
\setlength{\tabcolsep}{4pt}
\begin{tabular}{lrr}
\toprule
Split & Docs & Entities \\
\midrule
Train         & 287{,}202 & 388{,}422 \\
Dev           & 31{,}911  & 43{,}412  \\
Test          & 13{,}444  & 17{,}134  \\
\bottomrule
\end{tabular}
\caption{Per-split statistics for the proprietary clinical NER task.}
\label{tab:realworld-ner-stats}
\end{table}

\paragraph{Evaluation protocol.}
Each model is finetuned on the train split and evaluated under 3 seeds on the held-out test split.

\section{FineMed and FineMed-rephrased Construction}
\label{app:finemed-compute}

\subsection{Per-Source Corpus Statistics}
\label{app:per-source-stats}

Table~\ref{tab:dataset-stats-per-source} breaks down \emph{FineMed} variants by source (FineWeb-2, FinePDFs, FineWiki).

\begin{table}[h]
\centering
\scriptsize
\setlength{\tabcolsep}{3pt}
\begin{tabular}{lrrrrr}
\toprule
Corpus & Docs & Words & Length & Edu & Density \\
\midrule
\textbf{FineMed}              & \textbf{21.1M} & \textbf{19.2B} & \textbf{369} & \textbf{2.09} & \textbf{0.079} \\
\quad FineWeb-2               & 18.9M & 12.0B & 346  & 2.01 & 0.080 \\
\quad FinePDFs                & 2.1M  & 7.2B  & 766  & 2.76 & 0.069 \\
\quad FineWiki                & 38.6k & 26.5M & 283  & 3.12 & 0.140 \\
\midrule
\textbf{FineMed-filtered}     & \textbf{2.1M}  & \textbf{3.8B}  & \textbf{665} & \textbf{4.37} & \textbf{0.198} \\
\quad FineWeb-2               & 1.7M  & 2.0B  & 601  & 4.34 & 0.197 \\
\quad FinePDFs                & 322k  & 1.8B  & 1656 & 4.56 & 0.200 \\
\quad FineWiki                & 13.4k & 11.6M & 383  & 4.46 & 0.269 \\
\midrule
\textbf{FineMed-rephrased}    & \textbf{13.6M} & \textbf{4.5B}  & \textbf{191} & \textbf{2.86} & \textbf{0.164} \\
\quad FineWeb-2               & 11.9M & 3.0B  & 173  & 2.77 & 0.167 \\
\quad FinePDFs                & 1.7M  & 1.6B  & 547  & 3.46 & 0.143 \\
\quad FineWiki                & 26.1k & 10.1M & 212  & 3.51 & 0.243 \\
\bottomrule
\end{tabular}
\caption{Per-source breakdown of \emph{FineMed} variants, with high-level totals (bold) and per-source sub-rows.}
\label{tab:dataset-stats-per-source}
\end{table}

\subsection{Annotator Inference}
\label{app:axis-compute}

Table~\ref{tab:axis-compute} reports H100 GPU-hours for each annotation axis, broken down into the LLM-annotation stages and per-source student inference. Classifier training runs at 4--7 GPU-h per axis.

\paragraph{LLM annotation.}
We use vLLM~\citep{kwon2023efficient} with $\mathrm{TP}{=}4$ on H100s: Qwen3-30B-A3B-Instruct in bf16, Qwen3-235B-A22B-Instruct as the native-FP8 checkpoint. We report annotation hours as $4 \times$ wall-clock.

\paragraph{Student inference.}
Each axis runs on a single H100. ModernCamemBERT classifiers (subdomain, educational quality) use flash-attention-2 in bf16 with 8192-token input. GLiNER2 (medical-term density) runs in bf16 with score threshold 0.5 on the middle 512 tokens of each document (also the slice over which density is computed).

\begin{table*}[h]
\centering
\footnotesize
\setlength{\tabcolsep}{6pt}
\begin{tabular}{llrr}
\toprule
Step & Model & \#docs & GPU-hours \\
\midrule
\multicolumn{4}{l}{\emph{Subdomain classifier}} \\
LLM annotation, stage 1       & Qwen3-30B-A3B-Instruct      & 1M & 78.0 \\
LLM annotation, stage 2       & Qwen3-235B-A22B-Instruct & 500k     & 104.5 \\
Student inference (FineWeb-2) & ModernCamemBERT (finetuned) & 18.9M    & 5.9 \\
Student inference (FinePDFs)  & ModernCamemBERT (finetuned) & 2.1M     & 3.0 \\
Student inference (FineWiki)  & ModernCamemBERT (finetuned) & 38.6k    & 0.2 \\
\midrule
\multicolumn{4}{l}{\emph{Educational-quality scorer}} \\
LLM annotation, stage 1       & Qwen3-30B-A3B-Instruct      & 1M & 69.2 \\
LLM annotation, stage 2       & Qwen3-235B-A22B-Instruct & 100k     & 18.8 \\
Student inference (FineWeb-2) & ModernCamemBERT (finetuned) & 18.9M    & 4.8 \\
Student inference (FinePDFs)  & ModernCamemBERT (finetuned) & 2.1M     & 3.1 \\
Student inference (FineWiki)  & ModernCamemBERT (finetuned) & 38.6k    & 0.1 \\
\midrule
\multicolumn{4}{l}{\emph{Medical-term-density extractor}} \\
LLM annotation, extract pass  & Qwen3-235B-A22B-Instruct & 300k     & 29.5 \\
LLM annotation, review pass   & Qwen3-235B-A22B-Instruct & 300k     & 78.9 \\
Student inference (FineWeb-2) & GLiNER2 (finetuned)         & 18.9M    & 76.4 \\
Student inference (FinePDFs)  & GLiNER2 (finetuned)         & 2.1M     & 15.8 \\
Student inference (FineWiki)  & GLiNER2 (finetuned)         & 38.6k    & 0.3 \\
\bottomrule
\end{tabular}
\caption{H100 GPU-hours for the multi-axis annotation pipeline. \emph{\#docs} is the number of documents processed at that step (LLM-annotated sample sizes for the LLM rows; full retained-corpus sizes for the student-inference rows).}
\label{tab:axis-compute}
\end{table*}

\subsection{Rephrasing and Re-Annotation}
\label{app:rephrasing-inference}

We run the signal-amplifying rephrasing recipe on 4 H100 GPUs per task with vLLM~\citep{kwon2023efficient}. Because rephrasing changes per-document educational-quality and medical-term-density distributions, we re-run the educational-quality scorer and density extractor on the rephrased text, each on a single H100 with the configuration from \S\ref{app:axis-compute}. Table~\ref{tab:rephrasing-compute} reports per-source GPU-hours.

\begin{table}[h]
\centering
\footnotesize
\setlength{\tabcolsep}{6pt}
\begin{tabular}{lrrr}
\toprule
Step & FW2 & FinePDFs & FineWiki \\
\midrule
LLM rephrasing          & 4058 & 876  & 6   \\
Med-term re-extraction  & 48.7 & 10.1 & 0.2 \\
Edu rescoring           & 28.6 & 21.0 & 0.3 \\
\bottomrule
\end{tabular}
\caption{H100 GPU-hours for signal-amplifying rephrasing and the subsequent re-annotation of rephrased text. Rephrasing uses 4 H100s per job; re-annotation uses 1 H100.}
\label{tab:rephrasing-compute}
\end{table}

\subsection{Rephrased-Output Post-Processing Stats}
\label{app:output-filtering-stats}

We post-process rephrased outputs with DataTrove~\citep{penedo2024datatrove} (language identification then Gopher repetition filter; recipe in Appendix~\ref{app:rephrasing-postproc}). Table~\ref{tab:rephrasing-postproc-stats} reports per-source drop rates.

\begin{table}[h]
\centering
\footnotesize
\setlength{\tabcolsep}{6pt}
\begin{tabular}{lrrr}
\toprule
Corpus & Input docs & LangID & Gopher \\
\midrule
FineWeb-2  & 12.16M & 0.25\% & 2.21\% \\
FinePDFs   &  1.89M & 1.30\% & 6.99\% \\
FineWiki   &  27.1k & 0.92\% & 2.60\% \\
\bottomrule
\end{tabular}
\caption{Drop rates from rephrased-output post-processing, reported as a percentage of the step input.}
\label{tab:rephrasing-postproc-stats}
\end{table}

\subsection{Rephrasing Effect on Edu and Density Distributions}
\label{app:rephrasing-shift}

\begin{figure*}[h]
\centering
\includegraphics[width=0.95\textwidth]{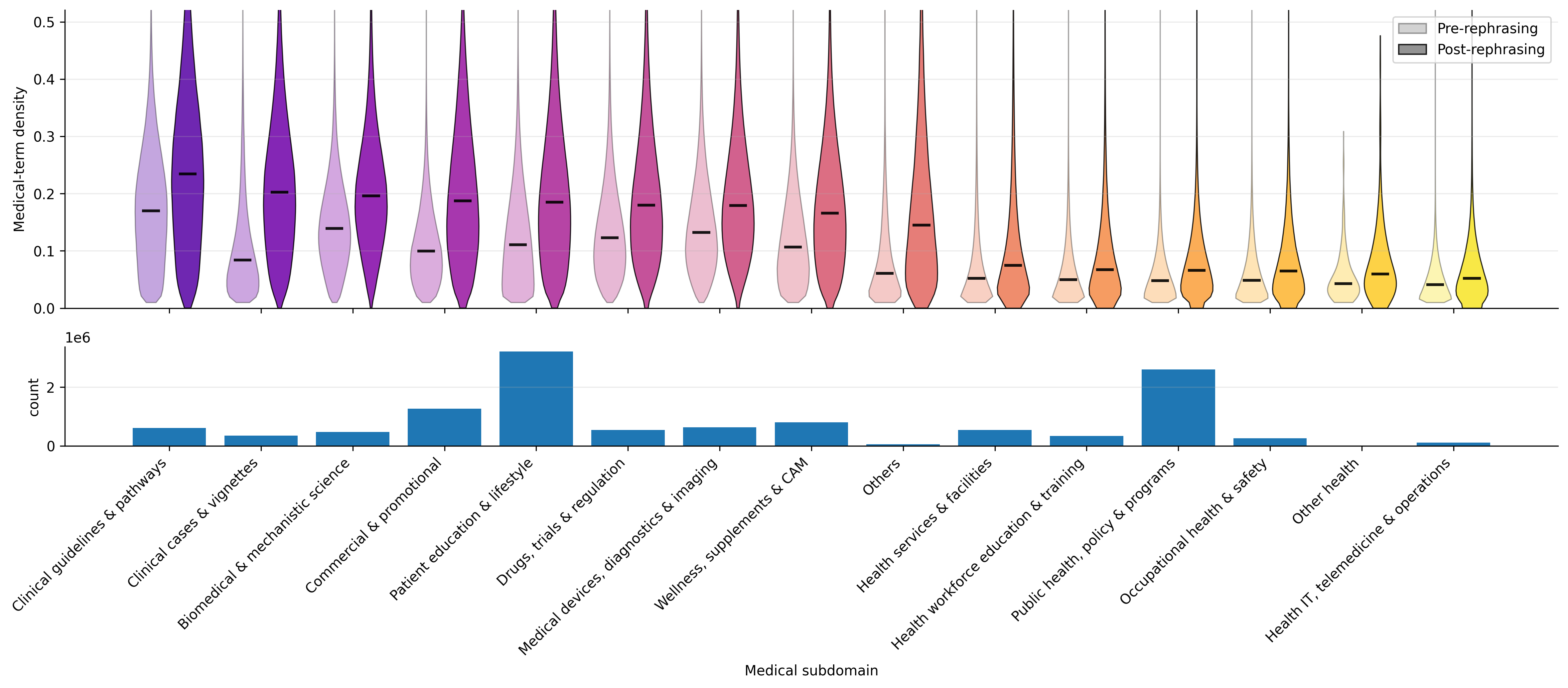}
\caption{Distribution of medical-term density per subdomain before (light) and after (dark) signal-amplifying rephrasing, on the FineWeb-2 medical subset. Bottom panel: document count per subdomain.}
\label{fig:rephrasing-shift-density}
\end{figure*}

\begin{figure*}[h]
\centering
\includegraphics[width=0.95\textwidth]{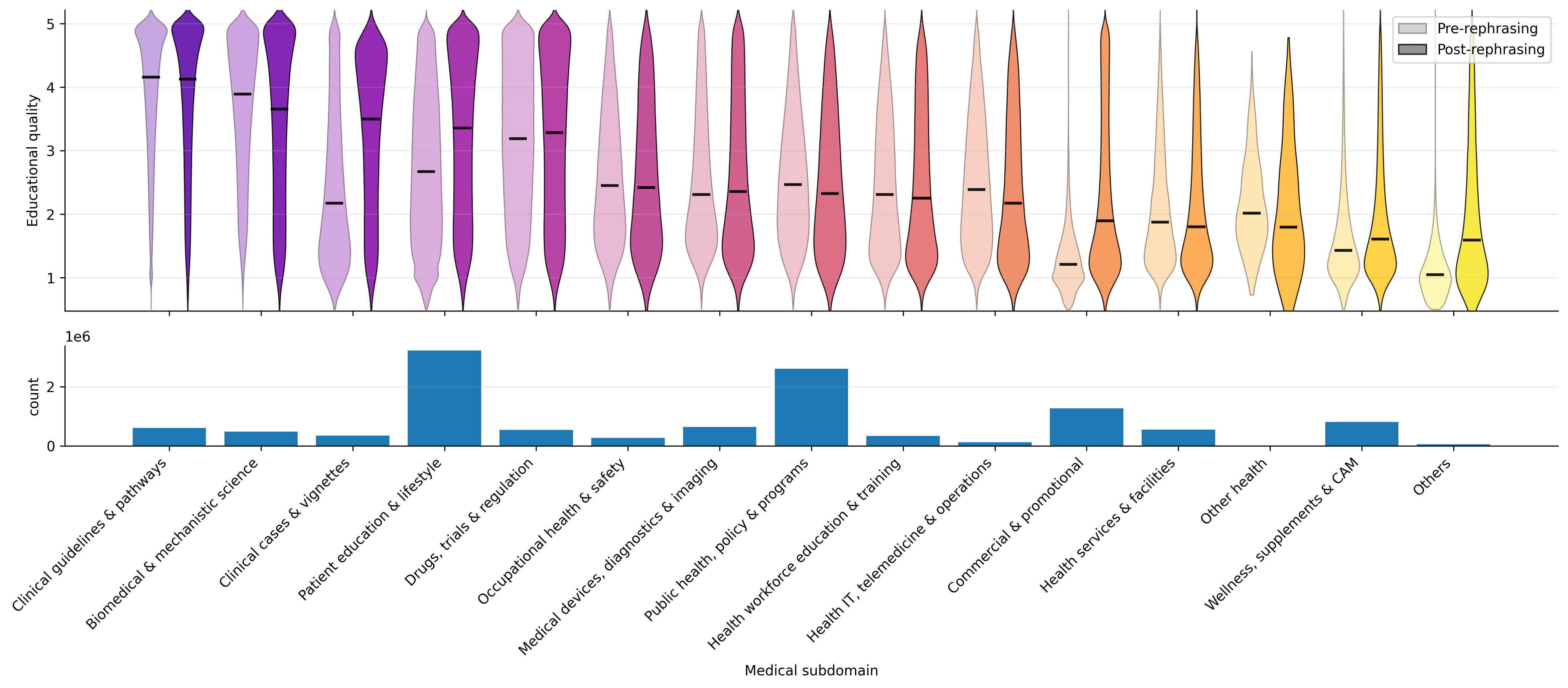}
\caption{Distribution of educational quality (0--5) per subdomain before (light) and after (dark) signal-amplifying rephrasing, on the FineWeb-2 medical subset. Bottom panel: document count per subdomain.}
\label{fig:rephrasing-shift-edu}
\end{figure*}

Rephrasing raises medical-term density across all 15 subdomains on the FineWeb-2 medical subset (Figure~\ref{fig:rephrasing-shift-density}), as expected from the faithful-densification design. Educational quality is not a direct target of the recipe; the per-subdomain shift on FineWeb-2 is mixed: some subdomains rise, others remain flat or drop slightly (Figure~\ref{fig:rephrasing-shift-edu}).

\section{Tokenizer}
\label{app:tokenizer}

\paragraph{Training.}
We train SentencePiece BPE on three filtered \emph{FineMed} sources: documents with $\geq 10$ words, subdomain $\neq$ \emph{Others}, edu $\geq 4$, and medical-term density $\geq 0.1$ (Table~\ref{tab:tokenizer-corpus}). The trained SentencePiece model is converted to a fast HuggingFace tokenizer with a final vocabulary of 50{,}368 (50{,}280 BPE pieces rounded to the next multiple of 64). Full configuration in Table~\ref{tab:tokenizer-config}.

\begin{table}[t]
\centering
\footnotesize
\setlength{\tabcolsep}{6pt}
\begin{tabular}{lrrr}
\toprule
Source     & Documents & Words & Lines \\
\midrule
FineWeb-2  & 1.82M     & 2.04B & 76.8M \\
FinePDFs   & 0.56M     & 1.75B & 138.8M \\
FineWiki   & 13.7k     & 11.6M & 0.57M \\
\midrule
Total      & 2.39M     & 3.81B & 216.2M \\
\bottomrule
\end{tabular}
\caption{Tokenizer training corpus, post-filter.}
\label{tab:tokenizer-corpus}
\end{table}

\begin{table}[t]
\centering
\footnotesize
\setlength{\tabcolsep}{4pt}
\renewcommand{\arraystretch}{1.15}
\begin{tabular}{ll}
\toprule
Parameter & Value \\
\midrule
Character coverage          & 0.9995 (byte fallback) \\
\texttt{byte\_fallback}     & \texttt{True} \\
\texttt{split\_digits}      & \texttt{True} (\texttt{500} $\to$ \texttt{5}, \texttt{0}, \texttt{0}) \\
Reserved symbols            & 24 (whitespace, newline, tab) \\
Normalizer                  & \texttt{Precompiled} charsmap (NFKC) \\
Pre-tokenizer               & \texttt{Metaspace} with \texttt{split=False} \\
Decoder                     & Byte-fallback \\
BERT specials               & \texttt{[UNK]}, \texttt{[CLS]}, \texttt{[SEP]}, \texttt{[PAD]}, \texttt{[MASK]} \\
EOS / padding               & \texttt{<|endoftext|>}, \texttt{<|padding|>} \\
PHI placeholders            & 11 (e.g., \texttt{|||PATIENT\_NAME|||}) \\
Markdown headers            & \texttt{\#} $\times$ 1 to \texttt{\#} $\times$ 6 \\
\bottomrule
\end{tabular}
\caption{DoctoBERT tokenizer configuration.}
\label{tab:tokenizer-config}
\end{table}

\paragraph{Fertility analysis.}
Table~\ref{tab:tokenizer-fertility} reports average tokens-per-word over 18 DrBenchmark French-medical subsets. Our tokenizer is trained on heterogeneous web-sourced \emph{FineMed}, distinct from DrBenchmark's curated medical text. This leaves a small fertility gap (1.43 vs.\ 1.41); our tokenizer with a 50k vocabulary (following ModernBERT) partially compensates, at the cost of a larger embedding layer.

\begin{table}[t]
\centering
\footnotesize
\setlength{\tabcolsep}{6pt}
\begin{tabular}{lrr}
\toprule
Tokenizer                                       & Vocab size & Fertility \\
\midrule
\multicolumn{3}{l}{\emph{French generalist}} \\
ModernCamemBERT                                 & 32{,}768   & 1.53 \\
\midrule
\multicolumn{3}{l}{\emph{English medical}} \\
BioClinical-ModernBERT                          & 50{,}368   & 2.00 \\
\midrule
\multicolumn{3}{l}{\emph{French medical}} \\
TransBERT-bio-fr                                & 32{,}002   & 1.46 \\
DrBERT                                          & 32{,}005   & \textbf{1.41} \\
CamemBERT-bio                                   & 32{,}005   & 1.59 \\
\midrule
\multicolumn{3}{l}{\emph{Ours}} \\
DoctoBERT-fr                                    & 32{,}768   & 1.48 \\
DoctoModernBERT-fr                              & 50{,}368   & 1.43 \\
\bottomrule
\end{tabular}
\caption{Average tokens-per-word over 18 DrBenchmark French-medical subsets. Lower is better. CamemBERT-bio~\citep{touchent_camembertbio_2024} is continually pretrained from the generalist CamemBERT and inherits its tokenizer, hence its higher fertility on medical text.}
\label{tab:tokenizer-fertility}
\end{table}

\section{Pretraining}
\label{app:pretraining}

This appendix details the document preparation and training hyperparameters underlying the \S\ref{sec:exp} ablations and the final \emph{DoctoBERT} (\S\ref{sec:final-doctobert}). Pretraining uses MosaicML Composer.\footnote{\url{https://github.com/mosaicml/composer}}

\paragraph{Document chunking.}
Encoder context windows are bounded.
We split long documents into training chunks rather than truncating them, to preserve the full token budget for pretraining.
The split walks a priority-ordered separator cascade (paragraph breaks, line breaks, sentence boundaries, whitespace), then greedily re-packs neighbouring chunks up to the target size to minimize padding.
Per-member chunk sizes are 510 for \emph{DoctoBERT-fr}, 1022 for \emph{DoctoModernBERT-fr} P1 (both reserving 2 tokens for \texttt{[CLS]} and \texttt{[SEP]}), and 8192 for P2 and P3.

\paragraph{Length-aware downsample.}
The ModernBERT P2 stage extends the context window from 1024 to 8192 tokens over a short 20B-token training window, so the training mix needs to be rich in long documents.
We downsample the 8192-token chunked corpus to bias it that way: within each source, we drop chunks below 128 words, then take 20\% of the remainder, with the long-document fraction ($\geq 2700$ words) of the output capped at 80\%.
Because the downsample applies the same shrink ratio to each source, source-mixture proportions are preserved.

\paragraph{Pretraining-corpus sizes.}
Training-corpus sizes vary across members and phases because chunking, the length-aware downsample, and the Bio\&Cli subset restriction operate at different scales (Table~\ref{tab:corpus-sizes}).
\emph{DoctoBERT-fr} and ModernBERT P1 share the same source mix but produce different chunk counts because their chunk sizes differ (510 vs 1022 tokens); P2 shrinks to roughly 30\% of its source after the length-aware downsample; and P3 narrows further to the Bio\&Cli subdomain subset.

\begin{table}[t]
\centering
\footnotesize
\setlength{\tabcolsep}{6pt}
\begin{tabular}{lrr}
\toprule
Configuration & \#Docs & \#Words \\
\midrule
DoctoBERT-fr P1        & 37.14M & 8.32B \\
DoctoBERT-fr P2        & 12.33M & 3.03B \\
DoctoModernBERT-fr P1  & 23.67M & 8.32B \\
DoctoModernBERT-fr P2  & 2.33M  & 2.53B \\
DoctoModernBERT-fr P3  & 2.71M  & 3.03B \\
\bottomrule
\end{tabular}
\caption{Pretraining-corpus sizes.}
\label{tab:corpus-sizes}
\end{table}

\paragraph{Phase progression on DrBenchmark.}
Table~\ref{tab:phase-progression} reports per-task DrBenchmark scores at intermediate training phases for both \emph{DoctoBERT} members. The Bio\&Cli annealing phase (\emph{DoctoModernBERT-fr} P3 / \emph{DoctoBERT-fr} P2) lifts most tasks over the preceding configuration.

\begin{table*}[t]
\centering
\footnotesize
\setlength{\tabcolsep}{4pt}
\begin{tabular}{l r@{}l r@{}l r@{}l r@{}l r@{}l r@{}l r@{}l}
\toprule
 & \multicolumn{4}{c}{QUAERO} & \multicolumn{4}{c}{E3C} & \multicolumn{2}{c}{MORFITT} & \multicolumn{2}{c}{DEFT2021} & \multicolumn{2}{c}{DIAMED} \\
\cmidrule(lr){2-5} \cmidrule(lr){6-9} \cmidrule(lr){10-11} \cmidrule(lr){12-13} \cmidrule(lr){14-15}
Configuration & \multicolumn{2}{c}{EMEA} & \multicolumn{2}{c}{MEDLINE} & \multicolumn{2}{c}{CLIN.} & \multicolumn{2}{c}{TEMP.} & \multicolumn{2}{c}{CLS} & \multicolumn{2}{c}{NER} & \multicolumn{2}{c}{CLS} \\
\midrule
\multicolumn{15}{l}{\emph{DoctoModernBERT-fr}} \\
P1 (200B) & 63.31 & ${}_{\pm 1.36}$ & 58.99 & ${}_{\pm 0.54}$ & 58.05 & ${}_{\pm 0.68}$ & 83.73 & ${}_{\pm 0.54}$ & 71.90 & ${}_{\pm 0.94}$ & 62.01 & ${}_{\pm 0.41}$ & 69.29 & ${}_{\pm 1.39}$ \\
P1$+$P2 ($+$20B 8192 context) & 63.46 & ${}_{\pm 0.97}$ & 59.65 & ${}_{\pm 1.60}$ & 58.35 & ${}_{\pm 2.04}$ & 82.88 & ${}_{\pm 0.22}$ & 71.56 & ${}_{\pm 0.88}$ & 63.14 & ${}_{\pm 0.73}$ & 70.82 & ${}_{\pm 3.20}$ \\
P1$+$P2$+$P3 ($+$20B Bio\&Cli) & 65.71 & ${}_{\pm 0.51}$ & 59.65 & ${}_{\pm 0.40}$ & 59.62 & ${}_{\pm 0.57}$ & 84.06 & ${}_{\pm 0.62}$ & 71.87 & ${}_{\pm 0.92}$ & 63.81 & ${}_{\pm 0.63}$ & 71.60 & ${}_{\pm 4.14}$ \\
\midrule
\multicolumn{15}{l}{\emph{DoctoBERT-fr}} \\
P1 (500B) & 67.48 & ${}_{\pm 1.44}$ & 62.12 & ${}_{\pm 0.59}$ & 61.05 & ${}_{\pm 1.17}$ & 84.90 & ${}_{\pm 0.68}$ & 72.42 & ${}_{\pm 0.46}$ & 66.51 & ${}_{\pm 0.45}$ & 72.65 & ${}_{\pm 2.65}$ \\
P1$+$P2 ($+$200B Bio\&Cli) & 68.39 & ${}_{\pm 0.84}$ & 62.54 & ${}_{\pm 0.45}$ & 62.75 & ${}_{\pm 1.62}$ & 84.60 & ${}_{\pm 0.51}$ & 73.36 & ${}_{\pm 0.26}$ & 66.41 & ${}_{\pm 0.43}$ & 72.56 & ${}_{\pm 1.23}$ \\
\bottomrule
\end{tabular}
\caption{Phase progression on DrBenchmark: per-task scores at intermediate training phases for both \emph{DoctoBERT} members. Per-task cells are mean$\pm$std F1 on the test split. Final-model rows match the corresponding entries in Table~\ref{tab:final}.}
\label{tab:phase-progression}
\end{table*}

\paragraph{Hyperparameters.}
Table~\ref{tab:hp} reports the settings used. The \S\ref{sec:exp} ablations share one configuration where only the training corpus varies. The final model spans \emph{DoctoBERT-fr} (two phases P1, P2) and \emph{DoctoModernBERT-fr} (three phases P1, P2, P3).

\begin{table*}[t]
\centering
\renewcommand{\arraystretch}{1.15}
\resizebox{\textwidth}{!}{
\scriptsize
\setlength{\tabcolsep}{3pt}
\begin{tabular}{lccc}
\toprule
Hyperparameter & Ablation (\S\ref{sec:exp}) & DoctoBERT-fr (\S\ref{sec:final-doctobert}) & DoctoModernBERT-fr (\S\ref{sec:final-doctobert}) \\
\midrule
\multicolumn{4}{l}{\emph{Architecture}} \\
Layers / hidden / heads / FFN     & 22 / 768 / 12 / 1152 (no biases) & 12 / 768 / 12 / 3072 & 22 / 768 / 12 / 1152 (no biases) \\
Attention                          & RoPE, sliding window 128 & learned absolute & RoPE, sliding window 128 \\
Vocabulary                         & 32{,}064~\citep{labrak_drbert_2023} & 32{,}768 & 50{,}368 \\
Parameters total / backbone / embedding & 135M / 110M / 25M & 111M / 85M / 26M & 149M / 110M / 39M \\
\midrule
\multicolumn{4}{l}{\emph{Training}} \\
Context length                     & 1024 & 512 & 1024 (P1) / 8192 (P2, P3) \\
Total tokens                       & 20B (\S\ref{sec:exp-filtering}; \S\ref{sec:exp-rephrasing} 1M scale) / 1B (\S\ref{sec:exp-rephrasing} 100k scale) & 500B (P1) / 200B (P2) & 200B (P1) / 20B (P2) / 20B (P3) \\
MLM mask rate                      & 0.30 & 0.15 & 0.30 (P1, P2) / 0.15 (P3) \\
Optimizer                          & \multicolumn{3}{c}{Decoupled StableAdamW, $\beta_1{=}0.9$, $\beta_2{=}0.98$, $\epsilon{=}10^{-6}$, $\text{wd}{=}10^{-5}$} \\
Peak learning rate                 & $8\times10^{-4}$ & $5\times10^{-4}$ & $8\times10^{-4}$ (P1) / $3\times10^{-4}$ (P2, P3) \\
LR schedule                        & warmup-stable-decay & linear warmup-decay to zero ($\alpha_f{=}0$) & warmup-stable (P1) / constant (P2) / inverse-sqrt anneal (P3) \\
Warmup                             & 200M / 10M tokens & 10B tokens & 2B (P1) / 0 (P2, P3) tokens \\
Global batch size                  & 4608 & 5120 & 4608 (P1) / 576 (P2, P3) \\
Precision                          & \multicolumn{3}{c}{bf16 (mixed-precision)} \\
\midrule
\multicolumn{4}{l}{\emph{Compute}} \\
Hardware                           & 4 H100s & \multicolumn{2}{c}{16 H100s (4 nodes $\times$ 4)} \\
Wall-clock (GPU-hours)             & --- & 985 (P1) / 355 (P2) & 311 (P1) / 36 (P2) / 36 (P3) \\
\bottomrule
\end{tabular}
}
\caption{Pretraining hyperparameters.}
\label{tab:hp}
\end{table*}

\clearpage
\section{Rephrasing Examples}
\label{app:rephrasing-examples}

Examples~\ref{fig:rephrase-example-1} and~\ref{fig:rephrase-example-2} illustrate two rephrasing modes: a register shift with compression, and entity-cooccurrence enrichment via specialist terminology.

\begin{table*}[t]
\centering
\scriptsize
\setlength{\tabcolsep}{6pt}
\renewcommand{\arraystretch}{1.2}
\begin{tabular}{|p{0.47\textwidth}|p{0.47\textwidth}|}
\hline
\textbf{Original} & \textbf{Rephrased} \\
\hline
A peine je viens de naître que je leur en fais voir de toutes les couleurs...

Mais je sais qu'ils m'aiment et qu'ils m'aimeront toujours.

Je n'ai pas respiré quand je suis née car j'avais une atrésie choanale.

Si vous êtes comme moi, c'est un terme bien barbare auquel on ne comprend rien !!

En fait, pour faire simple,

je n'avais pas de trous au niveau des narines.

Les orifices internes étaient bouchés par du cartilage.

Et comme les nouveaux nés ne savent pas respirer par la bouche,

je ne pouvais pas respirer...

Pour mon rythme cardiaque,

tout vient du fait que j'ai de nombreuses malformations,

dont une grosse CIV qui me cause de l'hypertension pulmonaire.

Pareil, pour faire simple,

le sang va et vient dans tous les sens au niveau de mon coeur

et çà me fatigue énormément, je manque constament d'oxygène...

Tout çà est bien difficile, et pour moi, et pour mes parents...

Je suis placée en soins intensifs sous haute surveillance

et Maman n'a le droit de me prendre dans ses bras qu'au bout de quelques jours.

J'ai des tuyaux partout, çà bippe sans cesse...

Dans quel monde ai-je là débarqué ???
&
Né : atrésie choanale bilatérale. Absence de respiration nasale initiale. Obstruction des orifices internes par cartilage. Nouvelle-née : impossibilité respiration buccale.

Cœur : cardiopathie congénitale multiple. CIV (communication interventriculaire) importante. HTAP (hypertension artérielle pulmonaire) secondaire. Shunt gauche-droit. Fatigue extrême. Hypoxie constante.

Prise en charge : soins intensifs (USIN). Surveillance continue. Monitoring cardiaque et respiratoire. Alitement strict.

Restriction : contact mère-enfant différé (quelques jours). Tubes multiples (intubation, voies veineuses). Alarmes continues.

Contexte familial : détresse parents. \\
\hline
\end{tabular}
\refstepcounter{rephraseexample}
\begin{center}Example~\therephraseexample. Personal narrative rewritten as a specialist clinical note: medical content preserved; lay paraphrases are replaced by acronyms of the same conditions (CIV, HTAP, USIN).\label{fig:rephrase-example-1}\end{center}
\end{table*}

\begin{table*}[t]
\centering
\scriptsize
\setlength{\tabcolsep}{6pt}
\renewcommand{\arraystretch}{1.2}
\begin{tabular}{|p{0.47\textwidth}|p{0.47\textwidth}|}
\hline
\textbf{Original} & \textbf{Rephrased} \\
\hline
Bonjour, c'est encore moi o/

Mon rhume est descendu (le coquin) est j'ai une n'en bronchite (comme d'hab...). Je fais des bronchites chronique (asthmatiforme sinon c'est pas drôle) à toux grasse (sinon c'est encore moins drôle).

Dans un de mes livres (je ne sais pas si j'en peux citer le titre mais c'est un livre très connu d'une dame :3) la synergie pour la' bronchite chronique grasse c'est

HE Myrte rouge

HE Ravinstara

HE Inule odorante

HV Calophylle

(et prendre du cyprès).

Ce que je ne comprends pas c'est Que l'on retrouve du Radiata uniquement en toux sèches, pourtant il me semble avoir lu qu'il était efficace pour évacuer les sécrétions... du coup je suis un peu perdue.

Merci d'avance.

Patakess

Bonjour Patakess,

L'Eucalyptus Radiata est en effet préférable pour les toux grasses, de par ses propriétés expectorantes et mucolytiques. Toutefois, comme elle est également anti-inflammatoire, elle peut donc avoir un rôle favorable dans la toux sèche.

En regardant l'ouvrage dont vous parlez, les deux synergies me paraissent appropriées, et c'est l'association des différentes huiles qui sera bénéfique, donc je pense que les deux synergies sont valables.

Sinon, vous pouvez essayer notre synergie Bronchite, qui comprend (entre autre) de l'Eucalyptus Radiata ?

En vous souhaitant une agréable journée.

Réponse de Patakess :

Bonjour.

Merci pour la réponse :). Étant donné que je n'ai pas de douleur, puis je remplacer le poivre noir de votre synergies par de la ravintsara ?

Encore merci :)

Réponse de Félicien CDS :

Bonjour,

Cela me parait tout à fait envisageable, et vous pouvez dans ce cas ajouter 30 gouttes de Ravintsara à la place des 15 de Poivre Noir. L'huile essentielle de Ravintsara étant immunostimulante, expectorante et anti-virale, elle renforcera l'action des autres huiles et vous permettra de récupérer plus rapidement.

En vous souhaitant une agréable journée.
&
Patiente : Patakess. Motif : exacerbation de bronchite chronique asthmatiforme à toux grasse. Antécédents : rhume récent ayant évolué vers bronchite chronique. Interrogatoire : absence de douleur thoracique. Demande : substitution du poivre noir (Piper nigrum) par l'huile essentielle de ravintsara (Cinnamomum camphora ct. 1,8-cinéole) dans la synergie bronchite.

Propriétés pharmacologiques de l'Eucalyptus Radiata : expectorant, mucolytique, anti-inflammatoire. Indication principale : toux grasse. Indication secondaire : toux sèche. Mécanisme d'action : évacuation des sécrétions bronchiques.

Synergie initiale proposée pour bronchite chronique grasse : huile essentielle de myrte rouge (Myrtus communis), huile essentielle de ravintsara (Cinnamomum camphora ct. 1,8-cinéole), huile essentielle d'inule odorante (Inula helenium), huile essentielle de calophylle (Calophyllum inophyllum), huile essentielle de cyprès (Cupressus sempervirens).

Modification posologique validée : remplacement de 15 gouttes de poivre noir par 30 gouttes d'huile essentielle de ravintsara. Justification : l'huile essentielle de ravintsara possède des propriétés immunostimulantes, expectorantes et antivirales. Effet attendu : renforcement de l'action des autres huiles essentielles, récupération clinique accélérée.

Références bibliographiques : ouvrage cité sur les synergies aromatiques pour bronchite chronique grasse. Site web de référence : Compagnie des Sens. \\
\hline
\end{tabular}
\refstepcounter{rephraseexample}
\begin{center}Example~\therephraseexample. Forum question rewritten as a pharmacist drug-information sheet: essential-oil entities preserved; context shifts to Latin botanical binomials and pharmacology register, expanding source terms without adding clinical facts.\label{fig:rephrase-example-2}\end{center}
\end{table*}

\clearpage
\section{Prompts}
\label{app:prompts}

This appendix collects the LLM prompts used for annotation (\S\ref{app:subdomain-classifier}, \S\ref{app:edu-classifier}, \S\ref{app:medterm-extractor}) and rephrasing (\S\ref{sec:rephrasing}).

\begin{figure*}[t]
\begin{tcolorbox}[colback=gray!8, colframe=gray!30, boxrule=0.4pt, arc=2pt, left=5pt, right=5pt, top=5pt, bottom=5pt]
\begin{lstlisting}[style=prompt]
You are an expert document classifier specializing in health-related content. Your task is to categorize web documents into exactly ONE TOPIC from a predefined list of health-related categories. Follow these instructions carefully to ensure accurate and consistent labeling.

Rules and Guidelines:

<guidelines>
1. Analyze the document: Carefully read the provided text to identify its primary focus, key themes, and specific terminology. If a URL is present, use it for context only; the document's text is the primary source of truth.
2. Select the best topic: Compare the document's content against the list of allowed topics and their definitions. Choose the one topic that most accurately reflects the document's main subject.
3. Construct reasoning: Write a concise justification for your topic selection. This reasoning must be 100 words or less and include 1-2 short, direct quotes from the text as evidence.
4. Handle exceptions: If the text is too short to analyze, is not clearly health-related, or consists mainly of navigational elements (like menus or footers), you must assign the topic "Others".
5. Strict topic selection: You must **choose exactly one topic** from the provided list. Do not invent new topics or alter the existing ones.
</guidelines>

Allowed Topics:

<topics>
... (15 classes; full names and descriptions in the taxonomy table above)
</topics>

Output Format:

Your response must be in strict JSON format with the following structure:

<output_format>
{
  "reasoning": "<A justification of 100 words or less, including 1-2 verbatim quotes.>",
  "topic": "<The single chosen topic name, exactly as provided in the list, without definitions.>"
}
</output_format>

Do not add any fields or wrap the JSON in markdown.
\end{lstlisting}
\end{tcolorbox}
\refstepcounter{prompt}
\begin{center}Prompt~\theprompt. Subdomain annotation prompt.\label{app:subdomain-prompt}\end{center}
\end{figure*}

\begin{figure*}[t]
\begin{tcolorbox}[colback=gray!8, colframe=gray!30, boxrule=0.4pt, arc=2pt, left=5pt, right=5pt, top=5pt, bottom=5pt]
\begin{lstlisting}[style=prompt]
Below is a proposed rubric to evaluate the quality of health-related content on
web pages. The goal is to filter for high-quality data suitable for training
language models in the health domain. Points are accumulated based on the
satisfaction of each criterion:

- Add 1 point if the extract provides basic health information that is clearly
focused on a medical or health topic. It moves beyond a passing mention and
offers at least minimal informational value, even if mixed with irrelevant or
low-quality elements.

- Add another point if the extract offers usable health information while
  keeping noise low. Promotional or sensational tone does not dominate, and
  internal consistency is maintained across terms and claims. This tier reflects
  baseline rigor without assessing external truth.

- Award a third point for specificity, density, and domain precision. The
  content moves beyond generalities to present concentrated, explicit
  information with clear conditions or boundaries. Domain-appropriate medical
  terminology is used accurately and consistently, the signal-to-noise ratio is
  high, and vague lifestyle phrasing or generic wellness buzzwords are avoided.
  Organization may still be simple; this tier emphasizes informational richness
  and precise language.

- Grant a fourth point if the extract is coherent and well-structured.
  Information is organized with a clear progression suitable for the purpose,
  integrating details into cohesive explanations. Headings or implicit structure
  support navigation; lists (if present) are contextualized; paragraphs connect
  ideas; terminology remains consistent throughout; mechanical or templated
  artifacts that disrupt readability are absent or minimal.

- Bestow a fifth point for expert synthesis or actionable guidance. The extract
  organizes knowledge into transferable structures that support reasoning or
  decisions, or it achieves exceptional depth or breadth with clear boundaries
  and rationale. Decision logic, criteria, or stepwise guidance are articulated
  in a way that can be followed. Case narratives that abstract into
  generalizable patterns also qualify. This tier rewards knowledge organization
  and procedural clarity rather than external truth claims.

After examining a given extract:

- Output in the following JSON format only (no extra text).
- Briefly justify your total score, up to 100 words.
- Conclude with an overall quality score as an integer from 0 to 5.

{
  "reasoning": "<Your justification, under 100 words>",
  "score": "<0-5 integer>"
}
\end{lstlisting}
\end{tcolorbox}
\refstepcounter{prompt}
\begin{center}Prompt~\theprompt. Educational-quality annotation prompt.\label{app:edu-prompt}\end{center}
\end{figure*}

\begin{figure*}[t]
\begin{tcolorbox}[colback=gray!8, colframe=gray!30, boxrule=0.4pt, arc=2pt, left=5pt, right=5pt, top=5pt, bottom=5pt]
\begin{lstlisting}[style=prompt]
## Objective

Extract **all** medical entities from the input text and classify each into one of the entity groups below.

## Extraction Rules

1. Strictly biomedical: Only extract entities with inherent **biomedical or clinical meaning**. A term qualifies only if its meaning is intrinsically medical — not merely because it appears in a clinical document.
2. Favor recall: Within the biomedical scope, if a term plausibly fits a group and is explicitly present, extract it.
3. Extract verbatim: Only extract text spans that appear exactly in the input. Do not infer, summarize, rephrase, or generate entities that are not explicitly present.
4. Longest span: Prefer the longest meaningful span (e.g., "acute myocardial infarction" over "infarction").
5. Preserve surface form: Keep exact case, punctuation, and spacing.
6. Include abbreviations: Extract medical abbreviations and acronyms (e.g., MI, COPD, MRI, CT).
7. Extract once: If an entity appears multiple times, include it only once per group.
8. One category per entity: Assign each entity to exactly one group.

## Extraction Order

- Process entity groups in this order: disease, drug, body_part, medical_procedure, molecular_marker, clinical_device, vital_function, living_beings.
- For each group, scan the text from beginning to end and output spans in the same order they appear.
- Omit any entity group with no matches.

## Exclusions

Do NOT extract:
- Ages, dates, durations: "63 years old", "2 weeks", "post-op day 3"
- Numeric values alone: "150 mL", "40%
- Person references: patient, doctor, child, mother, elderly, man, woman
- Facilities and locations: hospital, clinic, ICU, emergency room
- Administrative terms: referral, consultation, admission, discharge

## Entity Groups

<entity_groups>
... (8 classes; full names and descriptions in the taxonomy table above)
</entity_groups>

## Output Format

Return ONLY a JSON object. No explanations. No markdown code blocks.

{
   "entities": [
      {
         "entity": "<disease|drug|body_part|medical_procedure|molecular_marker|clinical_device|vital_function|living_beings>",
         "text": ["<exact span 1>", "<exact span 2>", ...]
      }
   ]
}

Only include entity groups that have at least one match. Omit empty groups.

## Quality Control

Before finalizing output, verify:
- All eligible entities are extracted and correctly categorized.
- Entity spans are exact and reflect the original text.
- No entity overlaps or duplications exist.
- Output JSON is valid with all required fields.
- Empty groups are omitted.
\end{lstlisting}
\end{tcolorbox}
\refstepcounter{prompt}
\begin{center}Prompt~\theprompt. Medical-entity extraction prompt (Pass 1).\label{app:medterm-prompt-pass1}\end{center}
\end{figure*}

\begin{figure*}[t]
\begin{tcolorbox}[colback=gray!8, colframe=gray!30, boxrule=0.4pt, arc=2pt, left=5pt, right=5pt, top=2pt, bottom=2pt]
\begin{lstlisting}[style=prompt]
## Objective

You are a medical entity extraction reviewer. Given a clinical text and an initial extraction of medical entities, review and correct the extraction by:
1. **Adding** any valid medical entities that were missed
2. **Removing** any false positives (non-medical terms or incorrectly categorized entities)
3. **Reclassifying** any entities assigned to the wrong category

## Review Guidelines

### What to ADD (Missed Entities)

Add entities that:
- Have clear biomedical or clinical meaning and appear verbatim in the text
- Are medical abbreviations or acronyms (e.g., MI, COPD, MRI, CT)
- Are drug names, disease names, anatomical terms, procedures, etc. that were overlooked
- Follow the longest span principle (e.g., "acute myocardial infarction" over "infarction")

### What to REMOVE (False Positives)

Remove entities that:
- Lack inherent biomedical meaning (general terms appearing in clinical context)
- Are demographic data: ages, dates, durations ("63 years old", "2 weeks")
- Are numeric values alone: "150 mL", "40%
- Are person references: patient, doctor, child, mother, elderly, man, woman
- Are facilities/locations: hospital, clinic, ICU, emergency room
- Are administrative terms: referral, consultation, admission, discharge
- Are personality traits or normal psychological states (e.g., perfectionism, anger)
- Are medical specialties (e.g., psychiatry, cardiology)
- Do NOT appear verbatim in the original text

### What to RECLASSIFY

Move entities to the correct category based on the definitions below.

## Entity Group Definitions

<entity_groups>
... (8 classes; full names and descriptions in the taxonomy table above)
</entity_groups>

## Review Process

1. **Analyze false positives**: Examine each entity in the initial extraction. Identify any that should be removed or reclassified.
2. **Identify missed entities**: Read the original text carefully. Find any medical entities not captured in the initial extraction.
3. **Reason through changes**: Document your reasoning for each modification.
4. **Produce corrected output**: Generate the final corrected extraction.

## Output Format

Return ONLY a JSON object with reasoning followed by the corrected extraction. No explanations outside JSON. No markdown code blocks.

{
   "reasoning": {
      "false_positives": "<List entities to remove and explain why each is not a valid medical entity or doesn't belong. Write 'None' if no false positives found.>",
      "reclassifications": "<List entities that need to move to a different category and explain why. Write 'None' if no reclassifications needed.>",
      "missed_entities": "<List entities that were missed in the initial extraction and explain why each should be added. Write 'None' if no missed entities found.>"
   },
   "entities": [
      {
         "entity": "<disease|drug|body_part|medical_procedure|molecular_marker|clinical_device|vital_function|living_beings>",
         "text": ["<exact span 1>", "<exact span 2>", ...]
      }
   ]
}

Notes:
- The "reasoning" section contains 3 free-form strings documenting your analysis.
- The "entities" section contains the final corrected extraction.
- Only include entity groups that have at least one match. Omit empty groups.
- Within each category, list entities in the order they first appear in the text.

## Quality Control

Before finalizing output, verify:
- All false positives from initial extraction are removed
- All missed entities are added
- All entities are correctly categorized
- Entity spans are exact and appear verbatim in the original text
- No duplicates exist within or across categories
- Output JSON is valid
- Empty entity groups are omitted
\end{lstlisting}
\end{tcolorbox}
\vspace{-8pt}
\refstepcounter{prompt}
\begin{center}Prompt~\theprompt. Medical-entity review prompt (Pass 2).\label{app:medterm-prompt-pass2}\end{center}
\end{figure*}

\begin{figure*}[t]
\begin{tcolorbox}[colback=gray!8, colframe=gray!30, boxrule=0.4pt, arc=2pt, left=5pt, right=5pt, top=5pt, bottom=5pt]
\begin{lstlisting}[style=prompt]
You are a Medical Corpus Engineer designing training data for a biomedical BERT model.

Given a source text, assess whether it contains medical or health information, then propose up to {n_stage1_pairs} reformulations as (genre, audience) pairs. `genre` and `audience` are free-form lowercase snake_case English identifiers — see suggested labels below; extend with new ones if needed, and reuse the same string for the same concept. The downstream pipeline uniformly samples one pair per document, so give the sampler a genuinely diverse menu.

Reasoning (generated first, ≤200 words) — MANDATORY STEPS:

Step 0 — Determine `has_medical_content` (1 sentence). If false, stop and return an empty `proposals` list.

Step 1 — Genre brainstorm (independent of audience): list 5–7 candidate genres the source's content could populate based ONLY on content sufficiency (entities, mechanisms, doses, findings, procedures), ignoring the source's framing. Format: `genre_name: realizable | not_realizable`.

Step 2 — Audience brainstorm (independent of genre): list 4–5 candidate audiences who would benefit from this content reframed for them, ignoring the source's natural reader. Format: `audience_name: appropriate | inappropriate`.

Step 3 — Couple {n_stage1_pairs} pairs from Step 1/Step 2 lists per the diversity rules below. Brainstorming dimensions independently first prevents the natural pull toward `(patient_education, patient_or_layperson)` from dragging both together.

Assessment:
A text has medical content if it discusses diseases, symptoms, treatments, anatomy, pharmacology, public health data, or clinical procedures with enough substance to support rewriting. Navigation menus, site descriptions, and pages with no medical facts should be marked as not having medical content.

## Suggested labels

Genres:
- Clinical documentation: `clinical_note`, `discharge_summary`, `case_report`, `consultation_letter`, `nursing_handover`, `referral_letter`, `medication_reconciliation`, `tumor_board_summary`
- Diagnostics: `diagnostic_report`, `lab_report`, `imaging_interpretation`, `pathology_report`, `prescription`
- Regulatory / pharmaceutical: `drug_information_sheet`, `patient_leaflet`, `clinical_guideline`, `informed_consent_form`, `clinical_protocol`
- Research / education: `research_abstract`, `systematic_review`, `patient_education`, `medical_qa`, `medical_textbook_section`
- Escape (last resort): `natural`

Audiences:
`medical_specialist`, `general_practitioner`, `nurse_or_allied_health`, `medical_student`, `researcher`, `patient_or_layperson`, `public_health_official`, `pharmacist`, `hospital_administrator`

## Realizability constraint (apply first)

A genre is realizable when the source contains enough medical facts to populate its structure — even when the source's tone or framing is completely different. A blog about migraine treatment can be realized as `clinical_note` (SOAP from symptoms/treatments), `drug_information_sheet` (the medications), or `medical_qa` (FAQ from the explanations) — not just `patient_education`. Do not let the source's natural style determine the choice. When no listed genre fits the content, choose `natural` rather than force-fitting.

## Diversity within realizability — applied during Step 3

If proposing multiple pairs, ALL of:

1. **Per-dimension breadth**: across your {n_stage1_pairs} proposals, use at least 3 distinct genres AND at least 3 distinct audiences.
2. **No near-duplicates**: if two proposals share both genre and audience, or could be summarized with the same one-sentence description, replace one. E.g., `(patient_education, layperson)` + `(patient_leaflet, layperson)` are duplicates.
3. **Anti-bias**: include `patient_education` in at most ONE proposal, and `patient_or_layperson` as audience in at most TWO.
4. **Pair-level plausibility**: each pair must be plausible in real medical writing. Do NOT couple `(prescription, researcher)`, `(research_abstract, layperson)`, or `(clinical_note, layperson)` — the brainstorm gave per-dimension flexibility; coupling must respect real-world combinations.
5. **Honesty floor**: if fewer than {n_stage1_pairs} pairs satisfy rules 1–4 from the realizable lists, return fewer pairs. Do NOT invent realizability.

## Language

`reasoning` uses natural prose. `genre` and `audience` are English machine identifiers (not translated). The stage-2 rewriter renders the document in {language}.
\end{lstlisting}
\end{tcolorbox}
\refstepcounter{prompt}
\begin{center}Prompt~\theprompt. Stage-1 rephrasing prompt (genre/audience proposal generation).\label{app:rephrase-stage1-prompt}\end{center}
\end{figure*}

\begin{figure*}[t]
\begin{tcolorbox}[colback=gray!8, colframe=gray!30, boxrule=0.4pt, arc=2pt, left=5pt, right=5pt, top=5pt, bottom=5pt]
\begin{lstlisting}[style=prompt, basicstyle=\tiny]
Rewrite the source text for BERT MLM training.

Target genre: `{genre}`
Intended audience: `{audience}`
Style: register=`{register}`, abbreviation=`{abbreviation}`

## Core Objective

Transform input text into **dense, medically informative output** optimized for BERT masked-language-model (MLM) pre-training.

BERT's MLM objective masks random tokens and predicts them from surrounding context. Training value is highest when:
* the masked token is a medical entity or medically-relevant word
* the surrounding context is informative (not generic filler)
* the same entities appear in varied local neighborhoods across the corpus

Your goal: make every output token count. Strip non-medical filler to raise entity density, and render the document in the specified genre/audience/register combination.

## Genre and audience escape clause

If the specified `{genre}` or `{audience}` cannot be realized from the source content without fabricating medical facts, either:
- reduce the scope to match what the source supports (omit sections the source does not fill)
- OR fall back to natural source organization if the genre does not apply at all

Never invent medical facts, values, entities, or diagnoses to satisfy a genre requirement.

## The Density-Volume Tradeoff (Critical)

* **DENSITY**: strip non-medical filler (boilerplate, ads, disclaimers, generic transitions) so that most output tokens carry medical information.
* **VOLUME**: preserve enough surrounding context for entities to appear in informative neighborhoods. Do not over-compress.

Rule of thumb: if the source has N medical facts, the output should retain all N facts with their contextual relationships — expressed more densely, not collapsed.

## Rewriting Principles

The goal is NOT to summarize or paraphrase — it is to distill text into its medically informative essence while preserving enough volume for effective BERT pre-training.

**Inclusion criterion**: Include content that teaches the model medical terminology, clinical relationships, or biomedical concepts. Strip content that fails this test (commercial metadata, navigation noise, generic transitions, geographic lists, pricing, legal disclaimers). Retain contextual language that situates medical entities — relational framing is the MLM learning signal.

### Output Formatting

Write in plain text. No markdown formatting (no `#`, `**`, ``` ``` ```, `---`) and no bullet-point lists (no `-`, `•`, `*` at the start of lines). When structure is needed, use numbered steps (`1.`, `2.`), labeled fields (`Label: value`), or prose. Labels must be in {language}.

### What to PRESERVE (Copy Exactly)
* Medical entities: diseases, symptoms, procedures, anatomical terms, devices
* Drug names (brand/generic), compounds, genes, proteins, species
* Domain idioms: clinical shorthand patterns, specialty-specific phrasings (even if informal)
* Quantitative data: doses, frequencies, lab values, measurements, ranges
* Codes: ICD, CPT, LOINC, NCT identifiers
* Negation scope: keep "no", "denies", "without", "ruled out" attached to their targets
* Uncertainty markers: "suspected", "possible", "likely", "probable"
* Temporal markers: dates, durations, relative time references

### What to STRIP
* Discourse filler: introductory transitions, concluding remarks, rhetorical hedges
* Commercial/distribution metadata: geographic availability, city/country lists, pricing, retailer names, OTC/prescription status by region
* Administrative noise: legal disclaimers, website navigation, social sharing prompts, cookie notices
* Non-clinical logistics: shipping information, purchase instructions, contact details for non-medical purposes

### You must NOT
* Invent specific values: do not fabricate lab results, vital signs, or dosages
* Add diagnoses not implied by the text
* Change negation polarity: "no fever" must not become "fever"
* Alter causal relationships
* Add medical entities, synonyms, or clarifications that are not in the source

### Placeholder & Anonymized Data Handling
Input text may contain placeholders or redacted fields due to privacy constraints (GDPR, HIPAA, etc.):
* **Non-medical PII** (names, addresses, dates of birth, ID numbers, contact info): Replace placeholders with fictional but realistic and **varied** values appropriate to the input language and locale. Avoid common/generic choices — use uncommon names, varied dates, diverse locations to prevent repetitive patterns across outputs.
* **Medical fields** (diagnoses, lab values, medications, vitals): Do NOT invent. If a medical value is redacted or missing, either omit that field entirely or mark it explicitly as missing/redacted.
* **Consistency**: If you generate a fictional value for a field, reuse that exact value for all occurrences of that field within the same output.

### Language Rule (Strict)
The rewritten text MUST be entirely in **{language}**.
* Use {language} medical register, terminology, and section headings
* Do not introduce foreign-language templates or headers
* Exception: Universally standardized abbreviations (DNA, RNA, mg, kg, ICD codes) may remain in their international form

## Style Dimension Reference

**register**: Linguistic formality level
- `formal`: Syntactically complete sentences with proper grammar, professional clinical register
- `telegraphic`: Fragmentary syntax, noun phrases over full clauses, minimal function words, note-like shorthand

**abbreviation**: Degree of lexical compression through shortened forms
- `expanded`: Predominantly full medical terminology, abbreviations used sparingly
- `moderate`: Balanced use of full terms and standard clinical abbreviations
- `heavy`: High density of shortened forms typical of time-pressured clinical documentation

Output ONLY the rewritten text in {language}. No preamble, no commentary.
\end{lstlisting}
\end{tcolorbox}
\refstepcounter{prompt}
\begin{center}Prompt~\theprompt. Stage-2 rephrasing prompt (rendering under sampled genre/audience).\label{app:rephrase-stage2-prompt}\end{center}
\end{figure*}

\end{document}